\documentclass[10pt,journal,compsoc]{IEEEtran}

\ifCLASSOPTIONcompsoc
  \usepackage[nocompress]{cite}
\else
  \usepackage{cite}
\fi
\usepackage{epsfig}
\usepackage{graphicx}
\usepackage{amsmath}
\usepackage{amssymb}
\usepackage{textcomp}
\usepackage{caption}
\usepackage{subcaption}
\usepackage[table]{xcolor}
\usepackage{makecell,multirow,diagbox}
\usepackage{setspace}
\usepackage{threeparttable}
\usepackage{float}
\graphicspath{{imgs/}}

\begin{document}
%
\title{GOT-10k: A Large High-Diversity Benchmark for Generic Object Tracking in the Wild}

\author{Lianghua~Huang,
        Xin~Zhao,~\IEEEmembership{Member,~IEEE,}
        and~Kaiqi~Huang,~\IEEEmembership{Senior Member,~IEEE}
\IEEEcompsocitemizethanks{\IEEEcompsocthanksitem L. Huang and X. Zhao are with the Center for Research on Intelligent System and Engineering, Institute of Automation, Chinese Academy of Sciences, Beijing 100190, China, and also with the University of Chinese Academy of Sciences, Beijing 100049, China. E-mail: huanglianghua2017@ia.ac.cn, xzhao@nlpr.ia.ac.cn.\protect\\
\IEEEcompsocthanksitem K. Huang is with the Center for Research on Intelligent System and Engineering and National Laboratory of Pattern Recognition, Institute of Automation, Chinese Academy of Sciences, Beijing 100190, China, and also with the University of Chinese Academy of Sciences, Beijing 100049, China, and the CAS Center for Excellence in Brain Science and Intelligence Technology, 100190 (kqhuang@nlpr.ia.ac.cn).}}
\IEEEtitleabstractindextext{%
\begin{abstract}
  We introduce here a large tracking database that offers an unprecedentedly wide coverage of common moving objects in the wild, called GOT-10k. Specifically, GOT-10k is built upon the backbone of WordNet structure~\cite{wordnet1995} and it populates the majority of over 560 classes of moving objects and 87 motion patterns, magnitudes wider than the most recent similar-scale counterparts~\cite{youtubebb2017,ilsvrc2015,datasetlasot2018,datasettrackingnet2018}. By releasing the large high-diversity database, we aim to provide a unified training and evaluation platform for the development of class-agnostic, generic purposed short-term trackers. The features of GOT-10k and the contributions of this paper are summarized in the following: (1) GOT-10k offers over 10,000 video segments with more than 1.5 million manually labeled bounding boxes, enabling unified training and stable evaluation of deep trackers. (2) GOT-10k is by far the first video trajectory dataset that uses the semantic hierarchy of WordNet to guide class population, which ensures a comprehensive and relatively unbiased coverage of diverse moving objects. (3) For the first time, GOT-10k introduces the \textit{one-shot} protocol for tracker evaluation, where the training and test classes are \textit{zero-overlapped}. The protocol avoids biased evaluation results towards familiar objects and it promotes generalization in tracker development. (4) GOT-10k offers additional labels such as motion classes and object visible ratios, facilitating the development of motion-aware and occlusion-aware trackers. (5) We conduct extensive tracking experiments with 39 typical tracking algorithms and their variants on GOT-10k and analyze their results in this paper. (6) Finally, we develop a comprehensive platform for the tracking community that offers full-featured evaluation toolkits, an online evaluation server, and a responsive leaderboard. The annotations of GOT-10k's test data are kept private to avoid tuning parameters on it. The database, toolkits, evaluation server and baseline results are available at http://got-10k.aitestunion.com.
\end{abstract}

}

\maketitle

\IEEEdisplaynontitleabstractindextext

\ifCLASSOPTIONpeerreview
\fi
%
\IEEEpeerreviewmaketitle

\IEEEraisesectionheading{\section{Introduction}\label{sec:introduction}}

\begin{figure*}[!ht]
  \centering
  \includegraphics[width=175mm]{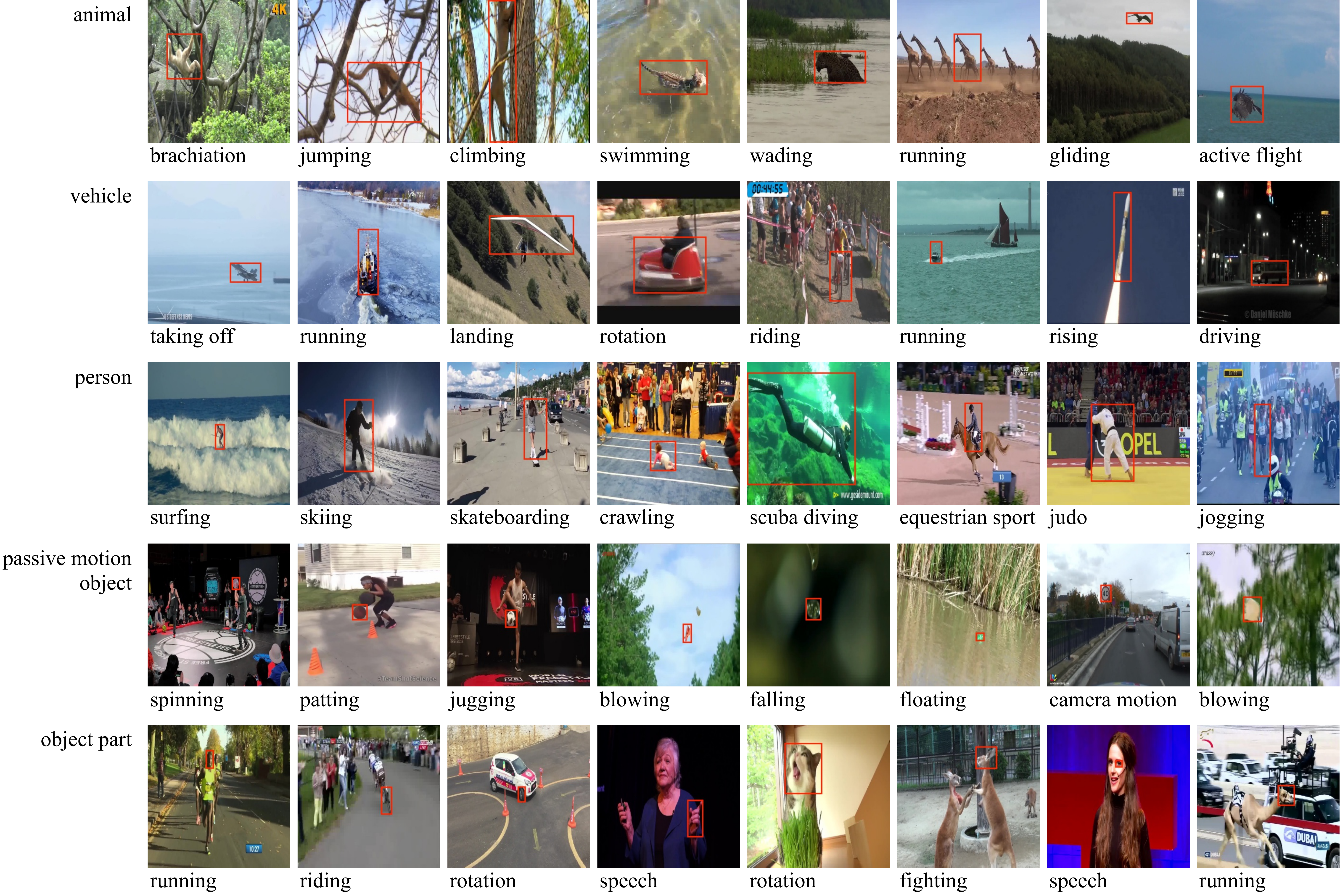}
  \caption{Screenshots of some representative videos collected and annotated in GOT-10k. Each video is attached with two semantic labels: object and motion classes. The object classes in GOT-10k are populated based on the semantic hierarchy of WordNet~\cite{wordnet1995}, where we expand five subtrees: \textit{animal, vehicle, person, passive motion object} and \textit{object part} to cover the majority of both natural and artificial moving objects in real-world. The motion classes are mostly backboned by WordNet, with only 6 exceptions: \textit{camera motion, dragon-lion dance, parkour, powered cartwheeling, taking off} and \textit{uneven bars}, which are not included in WordNet. GOT-10k populates 563 object classes and 87 motion classes in total. From the screenshots, we can also find that introducing motion labels to data collection significantly improves the variety of our dataset.}
  \label{fig:object_motion}
\end{figure*}
\begin{table*}[!ht]
\begin{center}
  \scriptsize
  \caption{Comparison of GOT-10k against other visual tracking and video object detection datasets in terms of scale, diversity, experiment setting, and other properties. GOT-10k is much larger than most datasets and it offers magnitudes wider coverage of object classes. Furthermore, GOT-10k introduces the \textit{one-shot protocol} to avoid evaluation bias towards familiar object classes. This differs from LaSOT and TrackingNet, where the training and test classes are completely overlapped with close distributions, leading to biased evaluation results towards seen classes.}
  \label{tab:tracking_datasets}
  \begin{threeparttable}
  \begin{tabular}{|
    >{\raggedleft\arraybackslash} m{2cm} ||
    >{\centering\arraybackslash} m{0.65cm} |
    >{\centering\arraybackslash} m{0.65cm} |
    >{\centering\arraybackslash} m{0.65cm} |
    >{\centering\arraybackslash} m{0.65cm} |
    >{\centering\arraybackslash} m{0.65cm} |
    >{\centering\arraybackslash} m{0.65cm} |
    >{\centering\arraybackslash} m{0.65cm} |
    >{\centering\arraybackslash} m{0.65cm} |
    >{\centering\arraybackslash} m{0.65cm} |
    >{\centering\arraybackslash} m{2.1cm} |
    >{\centering\arraybackslash} m{1.65cm} |
    >{\centering\arraybackslash} m{1.2cm} |}
    \hline
    \multirow{2}*[-0.7em]{Dataset} & \multicolumn{3}{c|}{Total} & \multicolumn{3}{c|}{Train} & \multicolumn{3}{c|}{Test} & \multicolumn{3}{c|}{Properties} \\
    \cline{2-13}
    & Classes & Videos & Boxes & Classes & Videos & Boxes & Classes & Videos & Boxes & Exp. Setting & Min/Max/Avg. Duration (seconds) & Frame Rate \\
    \hline
    \hline
    \textbf{OTB2015}~\cite{otb2015} & 22 & 100 & 59 k & - & - & - & 22 & 100 & 59 k & casual & 2.4/129/20 & 30 fps \\
    \textbf{VOT2019}~\cite{vot2019} & 30 & 60 & 19.9 k & - & - & - & 30 & 60 & 19.9 k & casual & 1.4/50/11 & 30 fps \\
    \textbf{ALOV++}~\cite{alov3002014} & 59 & 314 & 16 k & - & - & - & 59 & 314 & 16 k & casual & 0.63/199/16 & 30 fps \\
    \textbf{NUS\_PRO}~\cite{nuspro2016} & 12 & 365 & 135 k & - & - & - & 12 & 365 & 135 k & casual & 4.9/168/12 & 30 fps \\
    \textbf{TColor128}~\cite{datasetvotcolor2015} & 27 & 129 & 55 k & - & - & - & 27 & 129 & 55 k & casual & 2.4/129/14 & 30 fps \\
    \textbf{NfS}~\cite{datasetvotspeed2017} & 33 & 100 & 38 k & - & - & - & 33 & 100 & 38 k & casual & 0.7/86/16 & 240 fps \\
    \textbf{UAV123}~\cite{datasetuav2016} & 9 & 123 & 113 k & - & - & - & 9 & 123 & 113 k & casual & 3.6/103/31 & 30 fps \\
    \textbf{UAV20L}~\cite{datasetuav2016} & 5 & 20 & 59 k & - & - & - & 5 & 20 & 59 k & casual & 57/184/75 & 30 fps \\
    \textbf{OxUvA}~\cite{datasetvotlongterm2018} & 22 & 366 & 155 k & - & - & - & 22 & 366 & 155 k & open + constrained & 30/1248/142 & 30 fps \\
    \textbf{LaSOT}~\cite{datasetlasot2018} & 70 & 1.4 k & 3.3 M & 70 & 1.1 k & 2.8 M & 70 & 280 & 685 k & fully overlapped & 33/380/84 & 30 fps \\
    \textbf{TrackingNet}~\cite{datasettrackingnet2018} & 21 & 31 k & 14 M\tnote{*} & 21 & 30 k & 14 M\tnote{*} & 21 & 511 & 226 k & fully overlapped & -/-/16 & 30 fps \\
    \hline
    \hline
    \textbf{MOT15}~\cite{mot2015} & 1 & 22 & 101 k & 1 & 11 & 43 k & 1 & 11 & 58 k & - & 3/225/45 & 2.5$\sim$30 fps \\
    \textbf{MOT16/17}~\cite{mot2016} & 5 & 14 & 293 k & 5 & 7 & 200 k & 5 & 7 & 93 k & - & 15/85/33 & 14$\sim$30 fps \\
    \textbf{KITTI}~\cite{kitti2012} & 4 & 50 & 59 k & 4 & 21 & - & 4 & 29 & - & - & - & 10 fps \\
    \textbf{ILSVRC-VID}~\cite{ilsvrc2015} & 30 & 5.4 k & 2.7 M & 30 & 5.4 k & 2.7 M & - & - & - & - & 0.2/183/11 & 30 fps \\
    \textbf{YT-BB}~\cite{youtubebb2017} & 23 & 380 k & 5.6 M & 23 & 380 k & 5.6 M & - & - & - & - & - & 1 fps \\
    \hline
    \hline
    \textbf{GOT-10k} & \textbf{563} & 10 k & 1.5 M & \textbf{480} & 9.34 k & 1.4 M & \textbf{84} & 420 & 56 k & \textbf{one-shot} & 0.4/148/15 & 10 fps \\
    \hline
  \end{tabular}
  \begin{tablenotes}
    \item[*] TrackingNet is manually labeled at 1 fps while its all other annotations are automatically generated using correlation filters based tracking.
  \end{tablenotes}
  \end{threeparttable}
\end{center}
\end{table*}
\IEEEPARstart{G}{eneric} object tracking refers to the task of sequentially locating a moving object in a video, without accessing to the prior knowledge about the object (e.g., the object class) as well as its surrounding environment~\cite{otb2015,vot2013,alov3002014}. The task is highly challenging not only because of the class-agnostic nature in its definition, but also due to the unpredictable appearance changes during tracking, such as occlusion, object deformation, and background distraction. In real life, generic object tracking has a wide range of applications in diverse fields~\cite{survey_siamese2018} such as in surveillance~\cite{mot_vot2017}, augmented reality~\cite{augmented_reality2015}, biology~\cite{biology2018} and robotics~\cite{robotics2016}. Moreover, generic object tracking requires very few supervision during the tracking process. By exploring this, recent advances~\cite{wang2015understanding,wang2017selfsup} have further shown its potential in actively mining the training samples in unlabeled videos, paving the way for the more automatic learning system.

By definition, a generic object tracker has two important qualities: (1) It applies to a wide range of object classes; and (2) It is class-agnostic, meaning the algorithm also works for unseen classes of objects. However, such qualities have not been fully explored in recent large-scale tracking benchmarks~\cite{datasetlasot2018,datasettrackingnet2018}. Table~\ref{tab:tracking_datasets} compares current public tracking datasets in terms of scale, diversity, experiment setting, and others. We mainly compare our work with two recent large-scale datasets: LaSOT~\cite{datasetlasot2018} and TrackingNet~\cite{datasettrackingnet2018}, since other datasets are of small scales with no available training set. Although both datasets provide unified large training and test data, their manually defined object classes (21 and 70 classes, respectively) may not be representative enough for diverse real-world moving objects. Besides, in both datasets, the training and test object classes are fully overlapped with close distribution, leading to biased evaluation results towards familiar object classes, and the performance on them could be hard to be generalized to a wide range of unseen objects.

This work presents a large unified tracking benchmark that solves the above problems by (1) Using WordNet~\cite{wordnet1995} as the backbone to guide a comprehensive and unbiased population of object classes, and (2) Following the \textit{one-shot protocol} to avoid biased evaluation results. We summarize the contributions of this work in the following.
\begin{figure*}[!ht]
  \centering
  \includegraphics[width=180mm]{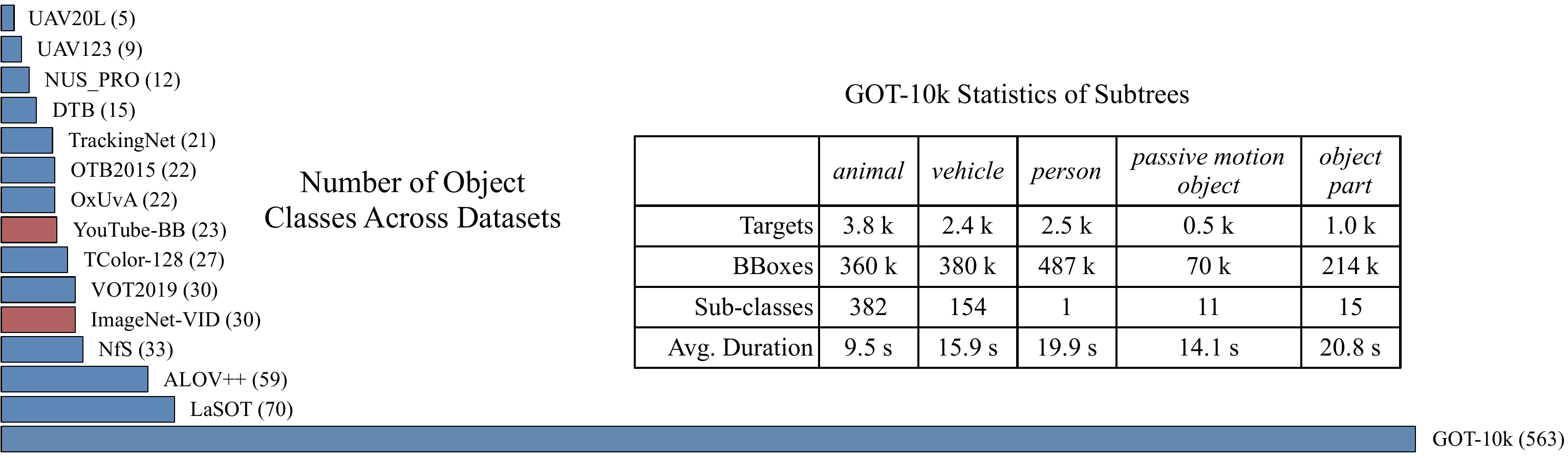}
  \caption{BAR CHART: Number of object classes in different tracking (blue) and video image detection (red) datasets. Among all the compared datasets, GOT-10k offers an unprecedentedly wider coverage of moving object classes. TABLE: Statistics of five subtrees populated in GOT-10k. All 563 object classes in GOT-10k are expanded from the five subtrees. The table shows the number of tracking targets, the scale of manually annotated object bounding boxes, sub-class counts, and average video length per each subtree.}
  \label{fig:class_numbers}
\end{figure*}
\begin{itemize}
  \item We build a large-scale database named GOT-10k for short-term generic object tracking, which collects over 10,000 video segments and manually annotates more than 1.5 million high-precision bounding boxes. The database enables unified training and fair comparison of deep trackers. Representative screenshots of GOT-10k are visualized in Figure~\ref{fig:object_motion}.
  \item GOT-10k populates 563 object classes and 87 motion forms from WordNet~\cite{wordnet1995}, offering magnitudes wider coverage of real-world moving objects than similar scale counterparts~\cite{datasetlasot2018,datasettrackingnet2018,youtubebb2017,ilsvrc2015}. The use of WordNet also gives the data in a way unbiased of the dataset creators.
  \item For the first time, we introduce the \textit{one-shot protocol} for tracker evaluation, where the classes between training and test sets are \textit{zero-overlapped}. Such protocol avoids the evaluation bias towards familiar objects and it promotes generalization in tracker development.
  \item GOT-10k offers additional labels such as motion classes, object visible ratios and absence indicators, facilitating the development of motion-aware and occlusion-aware trackers.
  \item We construct a large stable test set that consists of 420 videos belonging to 84 object classes and 31 motion classes. We show experimentally that enlarging the scale and diversity of test data significantly improves the reliability of the evaluation.
  \item We propose class-balanced metrics mAO and mSR for the evaluation of generic object trackers, which avoid the evaluation results to be dominated by the object classes of larger scales.
  \item We benchmark 39 recent state-of-the-art tracking approaches and their variants on GOT-10k and analyze their performance in this paper. We retrain all deep trackers on our training data to ensure a meaningful comparison.
  \item We carry out extensive experiments to study the impact of different aspects of training data on the performance of deep trackers.
  \item Finally, we build a website that offers full-featured toolkits, an online evaluation server and a responsive leaderboard to facilitate the tracking community. The annotations of our test data are kept private to avoid tuning parameters on it. The database, toolkits, evaluation server, and leaderboard can be found at http://got-10k.aitestunion.com.
\end{itemize}

\section{Related Work}\label{sec:related_work}
\begin{table}
  \scriptsize
  \begin{center}
      \centering
      \caption{Attribute annotations of different datasets. Compared to existing tracking and video object detection datasets, GOT-10k provides more comprehensive attribute annotations including object and motion classes as well as frame-level occlusion and absence labels. Furthermore, GOT-10k is the first tracking dataset that uses WordNet as the backbone for the collection of object and motion classes.
      }
      \label{tab:attributes}
      \begin{tabular}{|
          >{\raggedright\arraybackslash} m{1.7cm} |
          >{\centering\arraybackslash} m{0.7cm} |
          >{\centering\arraybackslash} m{0.75cm} |
          >{\centering\arraybackslash} m{1.1cm} |
          >{\centering\arraybackslash} m{0.95cm} |
          >{\centering\arraybackslash} m{1.1cm} |}
          \hline
          Datasets & Object Class Label & Motion Class Label & Frame Occlusion Label & Frame Absence Label & WordNet Backbone \\
          \hline
          \hline
          OTB2015 & & & & & \\
          VOT2019 & & & \checkmark & & \\
          NUS\_PRO & \checkmark & & \checkmark & & \\
          UAV123 & \checkmark & & & & \\
          NfS & \checkmark & & & & \\
          LaSOT & \checkmark & & & \checkmark & \\
          TrackingNet & \checkmark & & & & \\
          \hline
          \hline
          ImageNet-VID & \checkmark & & & \checkmark & \\
          YouTube-BB & \checkmark & & & \checkmark & \\
          \hline
          \hline
          \textbf{GOT-10k} & \checkmark & \checkmark & \checkmark & \checkmark & \checkmark \\
          \hline
      \end{tabular}
  \end{center}
\end{table}
We discuss in this section some of the datasets and benchmarks that are most related to GOT-10k.

\subsection{Evaluation Datasets for Tracking}\label{sec:evaluation_datasets_for_tracking}
Since 2013, a number of object tracking datasets have been proposed and served as unified platforms for tracker evaluation and comparison. The OTB~\cite{otb2013,otb2015}, ALOV++~\cite{alov3002014} and VOT~\cite{vot2013,vot2014,vot2015} datasets represent the initial attempts to unify the test data and performance measurements for generic object tracking. OTB collects 51 and 100 moving objects respectively from previous works in its first~\cite{otb2013} and second~\cite{otb2015} versions, while ALOV++~\cite{alov3002014} provides a larger pool of over 300 videos. VOT~\cite{vot2013,vot2016,vot2017} is an annual visual object tracking challenge held every year in conjunction with ICCV and ECCV workshops since 2013. Later on, several other datasets have been proposed targeting on solving specific issues. They include the large-scale people and rigid object tracking dataset NUS\_PRO~\cite{nuspro2016}, long-term aerial tracking dataset UAV123/UAV20L~\cite{datasetuav2016}, color tracking dataset TColor-128~\cite{datasetvotcolor2015}, long-term tracking dataset OxUvA~\cite{datasetvotlongterm2018}, thermal tracking datasets PTB-TIR~\cite{ptbtir2018} and VOT-TIR~\cite{vot2016}, RGBD tracking dataset PTB~\cite{datasetvotrgbd2013} and high frame-rate tracking dataset NfS~\cite{datasetvotspeed2017}. These datasets play an important role in boosting the development of tracking methods. However, these datasets are relatively small-scale and they only provide test data, which are not suitable for the unified training and evaluation of deep learning based tracking methods.

More recent datasets TrackingNet~\cite{datasettrackingnet2018} and LaSOT~\cite{datasetlasot2018} offer a scale that is on par with our dataset. TrackingNet chooses around 30 thousand videos from YouTube-BB~\cite{youtubebb2017} to form its training subset, and it collects another 511 videos with similar class distribution as its evaluation subset; while LaSOT collects and annotates 1.4 thousand videos manually. Despite their large scales, their manually defined object classes (21 and 70 object classes, respectively) may not be representative enough for diverse real-world moving objects. Besides, in both datasets, the object classes between training and test sets are fully overlapped with close distribution, where the evaluation results of deep trackers may be biased to these classes, and the performance on them cannot reflect the generalization ability of these trackers.
The comparison of GOT-10k with other tracking datasets in terms of scale, diversity, attribute annotations and others are shown in Table~\ref{tab:tracking_datasets}, Table~\ref{tab:attributes} and Figure~\ref{fig:class_numbers}. GOT-10k is larger than most tracking datasets and it offers a much wider coverage of object classes. Furthermore, it is the only benchmark that follows the \textit{one-shot} protocol in tracker evaluation to avoid the evaluation bias towards seen classes.

\subsection{Training Datasets for Tracking}
\begin{figure}[t]
  \centering
  \begin{subfigure}[b]{0.23\textwidth}
      \includegraphics[width=\textwidth]{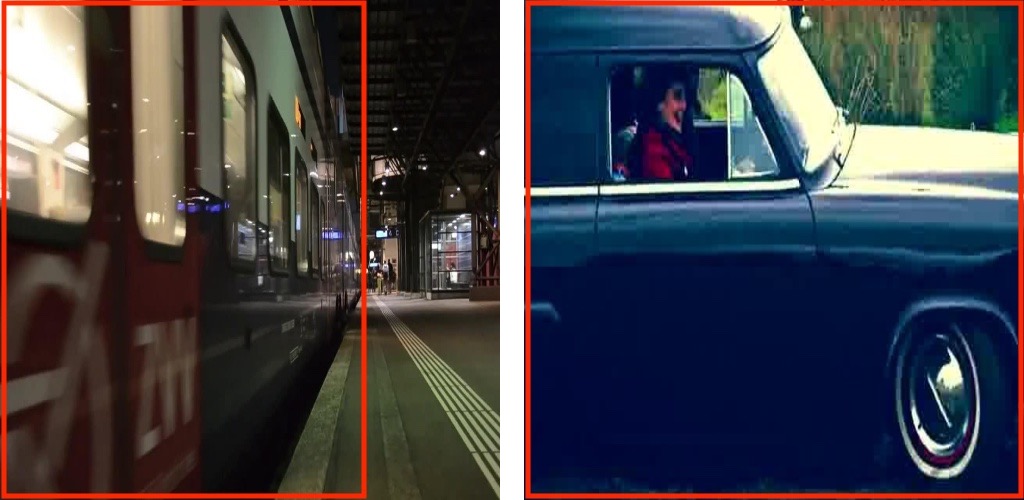}
      \caption{ImageNet-VID.}
  \end{subfigure}
  \begin{subfigure}[b]{0.23\textwidth}
      \includegraphics[width=\textwidth]{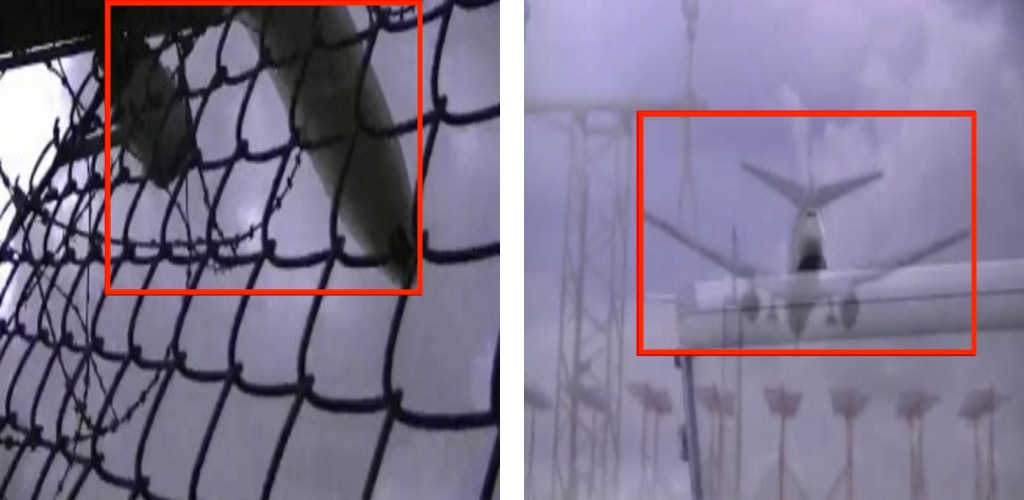}
      \caption{YouTube-BB.}
  \end{subfigure}
  \caption{Screenshots taken from the YouTube-BB~\cite{youtubebb2017} and ImageNet-VID~\cite{ilsvrc2015} datasets. Many of their videos contain noisy segments such as incomplete objects and shot changes, making them less optimal for the tracking task.}
  \label{fig:vid_screenshots}
\end{figure}

The arrival of the deep learning era has changed the paradigm of generic object tracking: Instead of starting training from scratch using a single annotated frame, deep learning based trackers usually learn some general knowledge (e.g., metrics, policies, and feature representation) from a large set of annotated videos, then transfer it to a specific target during tracking. Nevertheless, most traditional tracking datasets are small-scale and they only offer evaluation videos, which are not suitable for training deep trackers. Datasets that are mostly used for training trackers include ALOV++~\cite{alov3002014}, NUS\_PRO~\cite{nuspro2016}, ImageNet-VID~\cite{ilsvrc2015}, YouTube-BB~\cite{youtubebb2017}, LaSOT~\cite{datasetlasot2018} and TrackingNet~\cite{datasettrackingnet2018}. ImageNet-VID and YouTube-BB are video object detection datasets and they may contain noisy segments such as incomplete objects and shot changes (Figure~\ref{fig:vid_screenshots}), while other datasets are tracking datasets that always provide continuous long trajectories. Table~\ref{tab:tracking_datasets} compares GOT-10k with other video datasets in terms of scale and diversity. GOT-10k offers an annotation scale that is much larger than ALOV++ and NUS\_PRO and is on par with other large-scale datasets. Moreover, it provides magnitudes wider coverage of diverse object classes.

\section{Construction of GOT-10k}\label{sec:dataset_construction}

This section summarizes the technical details of the strategies and pipelines we use to build GOT-10k, shedding light on how the quality, coverage, and accuracy of GOT-10k are ensured. We also describe the experiments we carry out for the construction of a reliable test set.
\begin{figure*}[!ht]
  \centering
  \includegraphics[width=178mm]{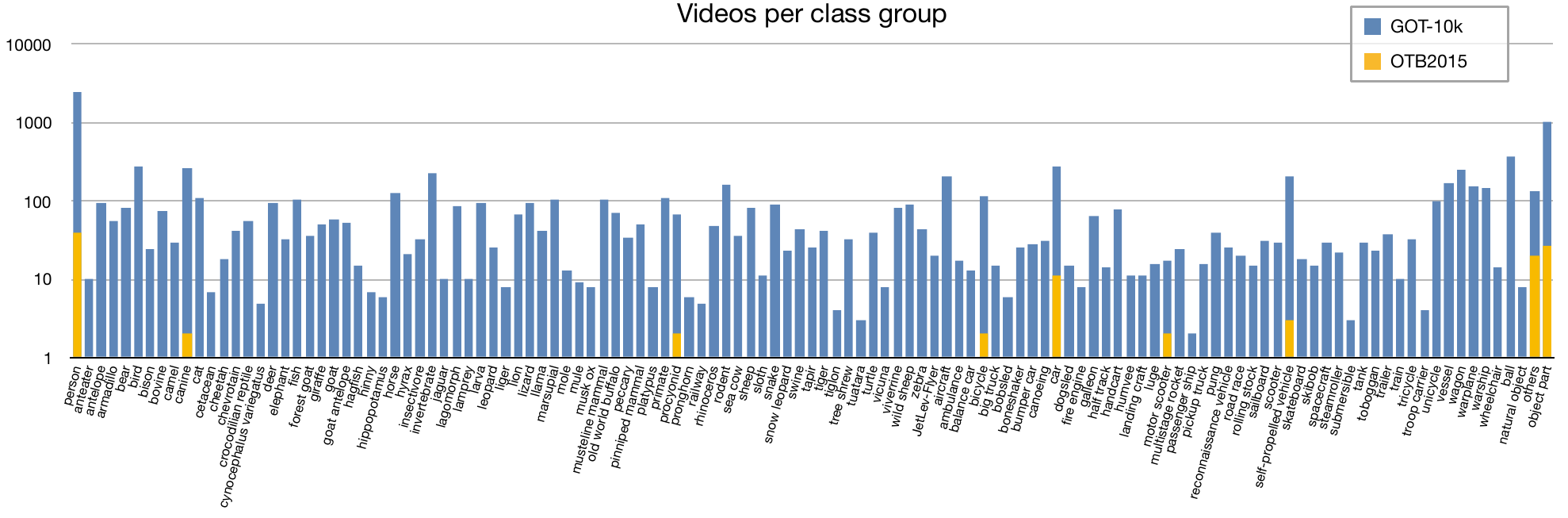}
  \caption{The number of videos per each group of object classes. In the collection stage, we categorize all potential object classes into 121 groups that we would like to ensure each being collected, as described in Section~\ref{sec:collection}. The plot shows the final distribution of these groups in GOT-10k, with OTB2015 dataset as a comparison.}
  \label{fig:distribution}
\end{figure*}
\subsection{Collection of Videos}\label{sec:collection}

The purpose of this work is to construct a large tracking dataset with a broad coverage of real-world moving objects, motion patterns, and scenes. To achieve such a comprehensive and unbiased coverage, we use WordNet~\cite{wordnet1995} as the backbone for the selection of object and motion classes. WordNet is a lexical database of English and it groups and organizes words (nouns, verbs, adjectives, and adverbs) based on their meanings. For example, nouns in WordNet are linked according to the \textit{hyponymy} relation (i.e., \textit{is a kind/member of}) to form a tree structure. The root node of nouns is \textit{entity} and it has several particulars such as \textit{object, thing, substance,} and \textit{location}.

Each collected video in GOT-10k is attached with two-dimensional labels: object and motion classes. We expand five nouns: \textit{animal, person, artifact, natural object,} and \textit{part} in WordNet to collect an initial pool of potential classes of moving objects, and we expand \textit{locomotion, action,} and \textit{sport} to collect motion classes. By manually filtering and pruning word subtrees (e.g., removing extinct, static and repeated object classes, grouping close sub-classes, etc.), we obtain a pool of around 2,500 object classes and over 100 motion classes. Although we can directly send these words to data collectors for video acquisition, the pool contains many uncommon object classes (e.g., \textit{broadtail, abrocome,} and \textit{popinjay}) making the collection process less efficient, and some common classes may even be overlooked. To improve the efficacy, we first categorize the 2,500 object classes into 121 groups (e.g., \textit{larva, canine, invertebrate,} and \textit{primate}) that we would like to ensure each being collected, then we rank the object classes in each group based on their corresponding searching volumes on the YouTube website over the last year. The searching volume reflects the popularity and the number of uploaded videos of each word, thus the ranking can guide collectors to find representative and qualified videos with a better chance.

We employ a qualified data company for video collection. The overall pipeline of video collection as well as verification is listed in Table~\ref{tab:qc_collection}. In summary, we carry out 1 collection and 5 verification stages to ensure the quality of each collected video. Defective videos that contain long-term stationary objects, noisy segments (e.g., shot changes, long-term target absence, too short or incomplete trajectories, etc.) or repeated scenarios are filtered out during the verification stages. For each collected data, the first 3 stages of verification are conducted in the company, then we have 2 trained verifiers to go through every video and determine whether to accept each or not. Afterwards, the authors of this work display all filtered videos and do the last screening (determining to accept each video or not). All rejected videos will have to be replaced by new ones collected by the data company. The verification process ensures that all the accepted videos have been visualized and checked by the authors and their quality can be guaranteed.

\begin{figure*}
  \centering
  \includegraphics[width=168mm]{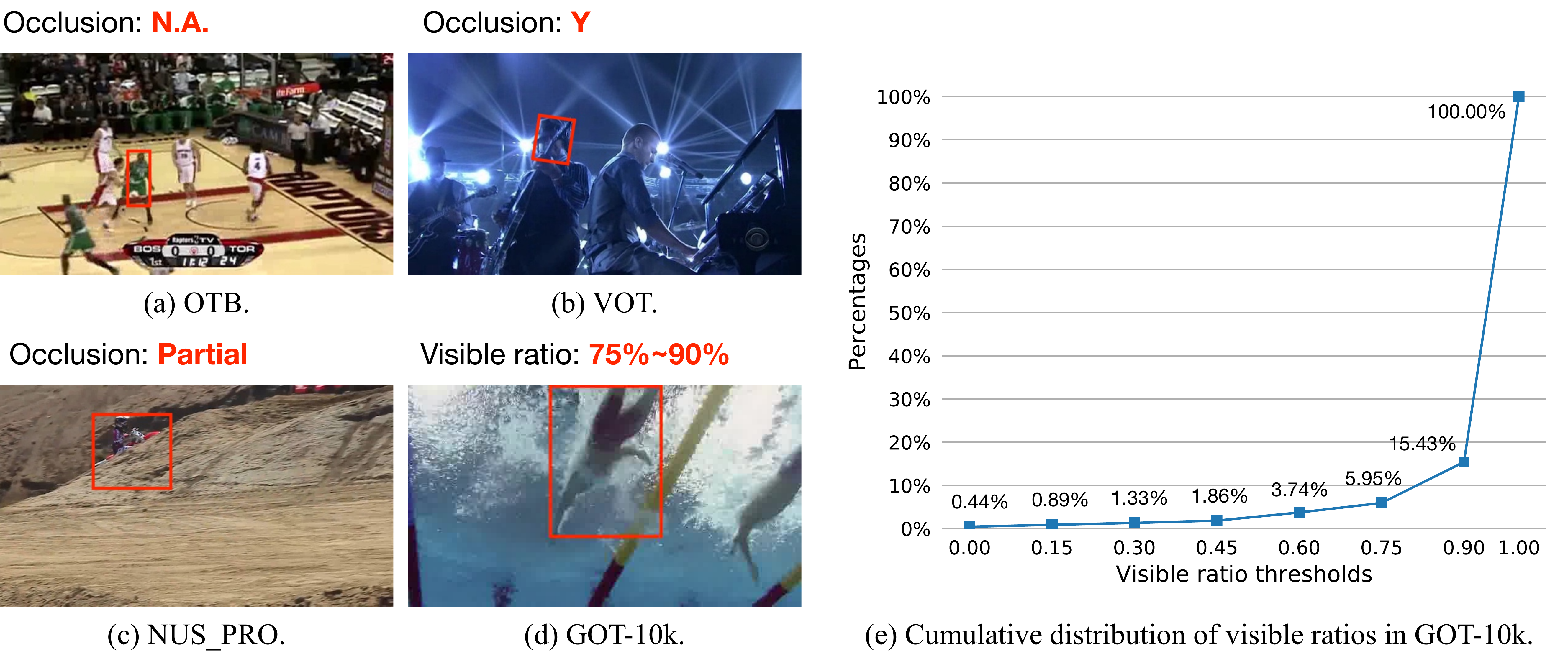}
  \caption{(a)-(d) Per-frame occlusion labeling in popular tracking datasets and GOT-10k. (a) The OTB dataset provides no frame-wise labeling of occlusion, while (b) The VOT dataset offers a binary label for each frame indicating whether or not the target is occluded. (c) The NUS\_PRO dataset distinguishes between partial and complete occlusions, while (d) GOT-10k further offers a more continuous labeling of target occlusion status. (e) Cumulative distribution of objects' visible ratios in GOT-10k. The regions that are occluded or truncated by image boundary correspond to the invisible parts. In around 15.43\% of frames, the targets are occluded/truncated (with less than 90\% visible), and in approximately 1.86\% of frames, they are heavily occluded/truncated (with less than 45\% visible). In about 0.43\% of frames, the targets are absent (being fully occluded or out-of-view).}
  \label{fig:occ_annotations}
\end{figure*}

The final pool contains 563 classes of moving objects and 87 classes of motion, with a total scale of around 10 thousand videos. Figure~\ref{fig:object_motion} illustrates some screenshots of GOT-10k videos labeled with varied object and motion classes. Figure~\ref{fig:distribution} shows the final distribution of video numbers of the 121 object class groups used in the collection stage, with OTB2015 dataset as a comparison.
From the figure, we observe that the distribution over different class groups is imbalanced. Such imbalance is mainly caused by the filtering process in the collection and verification stages. During data collection, we find that some object classes are rarely captured on YouTube with few qualified videos, while some others are much more frequently filmed with better quality and diversity. Also, during verification stages, videos of many classes are frequently filtered out due to their monotonous scenarios and motion, incomplete objects, very slow motion, or fragmented trajectories. After rounds of verification, the videos of our dataset naturally exhibit an imbalanced distribution over classes.

\begin{table}[t]
  \small
  \begin{center}
      \centering
      \caption{Quality control pipeline for video collection. The video collection as well as the first 3 stages of verification (marked with $*$) are conducted in the data company.}
      \label{tab:qc_collection}
      \begin{tabular}{|
          >{\centering\arraybackslash} m{0.7cm} |
          >{\centering\arraybackslash} m{2.2cm} |
          >{\centering\arraybackslash} m{2.5cm} |
          >{\centering\arraybackslash} m{1.8cm} |}
          \hline
          Stage & Description & Executor & Proportion \\
          \hline
          \hline
          $1^{*}$ & Data collection & Collectors & 100\% \\
          \hline
          $2^{*}$ & Verification & Collectors & 15\% $\scriptsize{\sim}$ 30\% \\
          \hline
          $3^{*}$ & Verification & Project team & 15\% $\scriptsize{\sim}$ 30\% \\
          \hline
          $4^{*}$ & Verification & Verification team & 5\% $\scriptsize{\sim}$ 10\% \\
          \hline
          $5$ & Verification & Our trained verifiers & 100\% \\
          \hline
          $6$ & Verification & The authors & 100\% \\
          \hline
      \end{tabular}
  \end{center}
\end{table}

Highly imbalanced class distribution is often observed in large-scale datasets such as ImageNet~\cite{imagenet2009} (10$\sim$2500 images per class), YouTube-BB~\cite{youtubebb2017} (26k$\sim$1.8M labels per class) and more recent Open Images~\cite{openimages2018} (10$\sim$10k labels per class). It is also a common phenomenon in the distributions of \textit{natural species} and \textit{Internet data} that is formally named \textbf{Long-Tailed Distribution}~\cite{longtail2015}. Long-tailed distribution is universal and learning under such imbalanced data is an important topic for real applicational usage, which has been studied over years~\cite{longtail2015,longtail2016,longtail2017,longtail2019}. We believe the class-imbalance of our training data casts meaningful challenges that encourage the design and development of more practical and scalable generic object trackers.

As a special case, similar to the YouTube-BB dataset, we give the \textit{person} class preferential treatment in terms of the total number of videos during data collection (which accounts for around 24\% of our entire database). The main reasons are two-fold. First, the \textit{person} class is especially important and both the research community and industry have a great interest in tracking people in videos. Second, \textit{person}s embrace a rich collection of appearance changes (e.g., diverse \textit{clothing, bags,} and \textit{accessories}) and motion forms (e.g., \textit{jogging, swimming, skiing, crawling, cycling, diving, climbing, equitation, judo, surfing, dancing,} and \textit{gymnastics}), and they appear in a variety of scenarios (e.g., \textit{forests, deserts, mountain cliffs, gymnasiums, parties,} and \textit{ballrooms}). Therefore, they tend to exhibit a much higher diversity than other classes of objects, which could be more advantageous for the learning of generic purposed trackers.

\subsection{Annotation of Trajectories}
\label{sec:annotation_of_trajectories}

We follow the standard in object detection~\cite{crowdsourcing2012} for the labeling of objects' tight bounding boxes. Note this differs from some visual tracking datasets such as VOT~\cite{vot2013,vot2016}, where the optimal object bounding box is defined as a rotated rectangle containing minimum background pixels. Since visual tracking algorithms have been frequently used in a number of related areas such as video object detection~\cite{tubelets2016,tcnn2017} and segmentation~\cite{seg2018}, multiple object tracking~\cite{mot_vot2017} and self-supervised learning~\cite{wang2015unsupervised,wang2017selfsup}, keeping a compatible annotation standard encourages the development of more practical trackers.

\begin{table}[t]
  \small
  \begin{center}
      \centering
      \caption{Quality control pipeline for trajectory annotation. The trajectory annotation as well as the first 3 stages of verification (marked with $*$) are conducted in the data company.}
      \label{tab:qc_annotation}
      \begin{tabular}{|
          >{\centering\arraybackslash} m{0.7cm} |
          >{\centering\arraybackslash} m{2.2cm} |
          >{\centering\arraybackslash} m{2.5cm} |
          >{\centering\arraybackslash} m{1.8cm} |}
          \hline
          Stage & Description & Executor & Proportion \\
          \hline
          \hline
          $1^{*}$ & Data annotation & Annotators & 100\% \\
          \hline
          $2^{*}$ & Verification & Annotators & 15\% $\scriptsize{\sim}$ 30\% \\
          \hline
          $3^{*}$ & Verification & Project team & 15\% $\scriptsize{\sim}$ 30\% \\
          \hline
          $4^{*}$ & Verification & Verification team & 5\% $\scriptsize{\sim}$ 10\% \\
          \hline
          $5$ & Verification & The authors & 100\% \\
          \hline
      \end{tabular}
  \end{center}
\end{table}

In addition to object bounding boxes, GOT-10k also provides the annotation of visible ratios. A visible ratio is a percentage indicating the approximate proportion of an object that is visible. The pixels that are occluded or truncated by image boundaries correspond to the invisible part. As is indicated in many tracking benchmarks~\cite{vot2017,otb2015,nuspro2016}, occlusion is one of the most challenging factors that can easily cause tracking failure. We hope the labeling of visible ratios can facilitate the tracking community to develop occlusion-aware tracking methods. We divide visible ratios into 7 ranges with a step of $15\%$. Figure~\ref{fig:occ_annotations} (a)-(d) compares the per-frame occlusion labeling of different tracking datasets while Figure~\ref{fig:occ_annotations} (e) shows the cumulative distribution of visible ratios annotated in our dataset. Through visible ratios, GOT-10k provides a more continuous labeling of target occlusion status (the percentage of occlusion/truncation can be easily calculated as $(1-v)$ using the visible ratio $v$). Similar to video collection, we use 1 annotation stage and 4 verification stages to ensure the quality of each annotation. Table~\ref{tab:qc_annotation} lists out the pipeline. For the annotated data, the first 3 stages of verification are conducted in the data company, then the authors will go through every video and determine whether to accept its annotations or not. All unqualified annotations are sent back to the company for refinement. The verification process ensures that all accepted videos have been visualized and checked for several rounds and their quality can be largely guaranteed.

\subsection{Dataset Splitting}\label{sec:dataset_splitting}

\begin{figure*}[!ht]
  \centering
  \begin{subfigure}[b]{0.245\textwidth}
      \includegraphics[width=\textwidth]{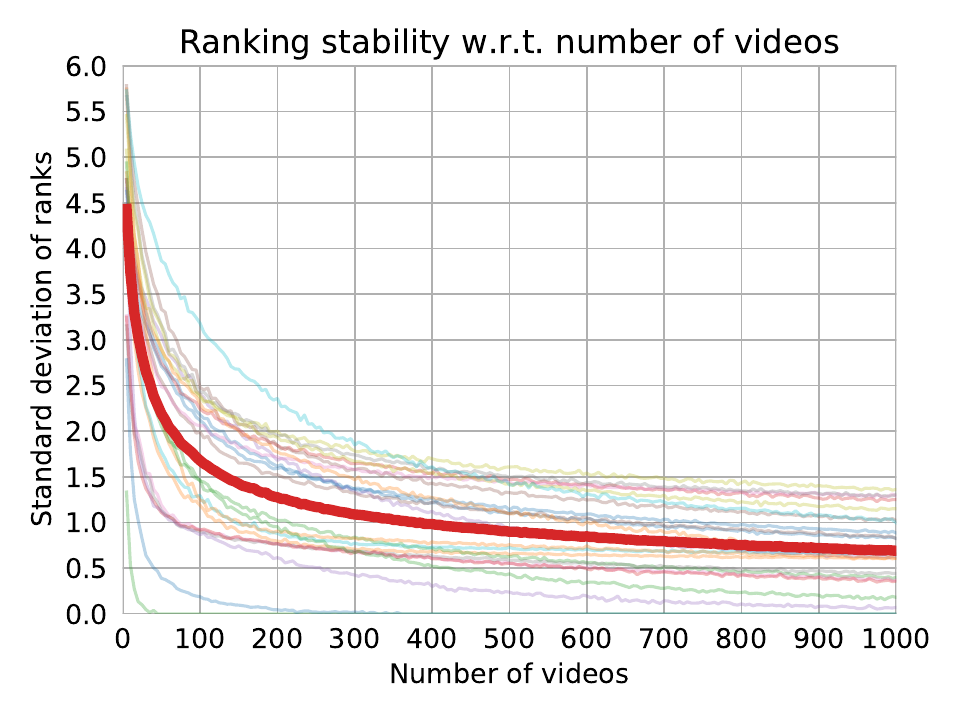}
      \caption{Impact of video number.}
      \label{fig:stdv_video_num}
  \end{subfigure}
  \begin{subfigure}[b]{0.245\textwidth}
      \includegraphics[width=\textwidth]{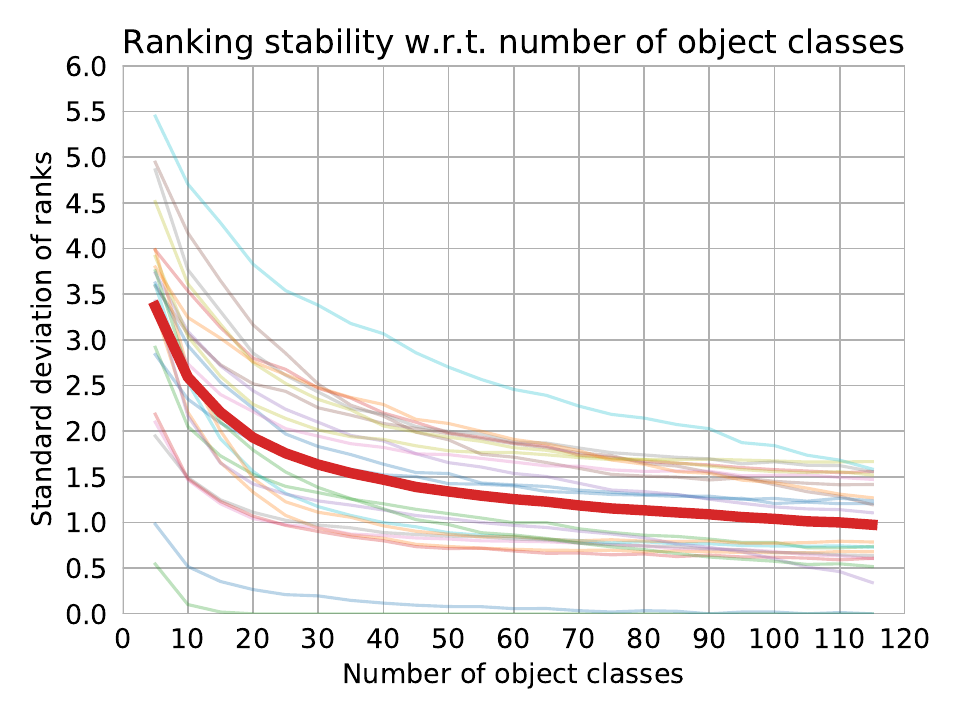}
      \caption{Impact of object classes.}
      \label{fig:stdv_obj_class}
  \end{subfigure}
  \begin{subfigure}[b]{0.245\textwidth}
    \includegraphics[width=\textwidth]{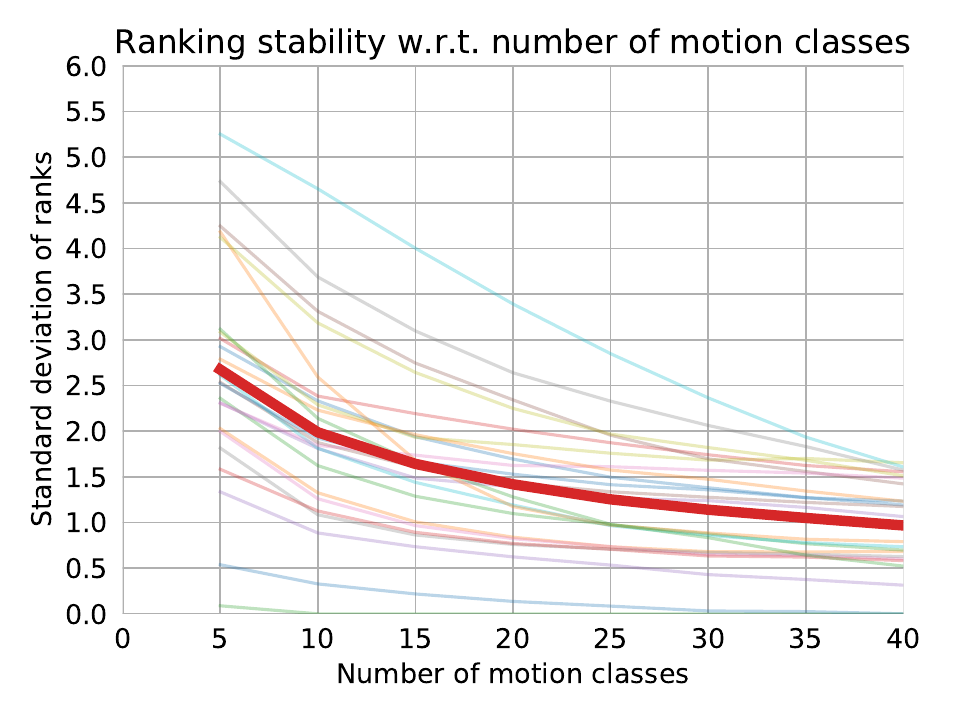}
    \caption{Impact of motion classes.}
    \label{fig:stdv_mot_class}
  \end{subfigure}
  \begin{subfigure}[b]{0.245\textwidth}
    \includegraphics[width=\textwidth]{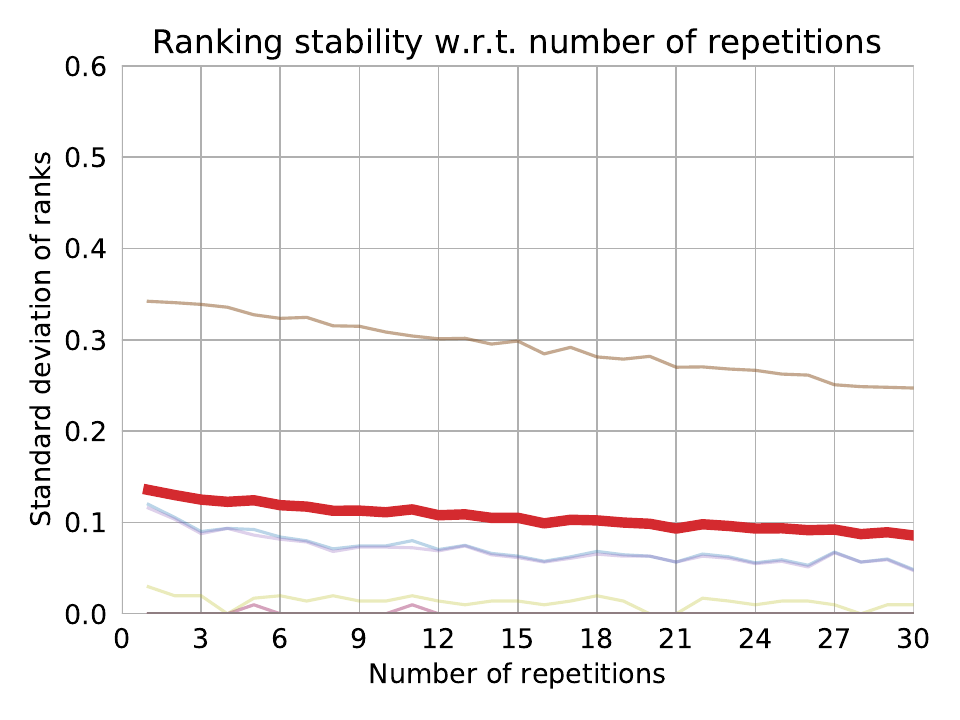}
    \caption{Impact of repetition time.}
    \label{fig:stdv_repetition}
  \end{subfigure}
  \caption{The impact of different test data configurations on evaluation stability, assessed by the standard deviation of the ranking of 25 trackers. The average standard deviation over trackers is emphasized by the thick red curves. A higher standard deviation indicates a less reliable evaluation. Better viewed with zooming in.}
  \label{fig:stdv}
\end{figure*}
\begin{figure}
  \centering
  \includegraphics[width=78mm]{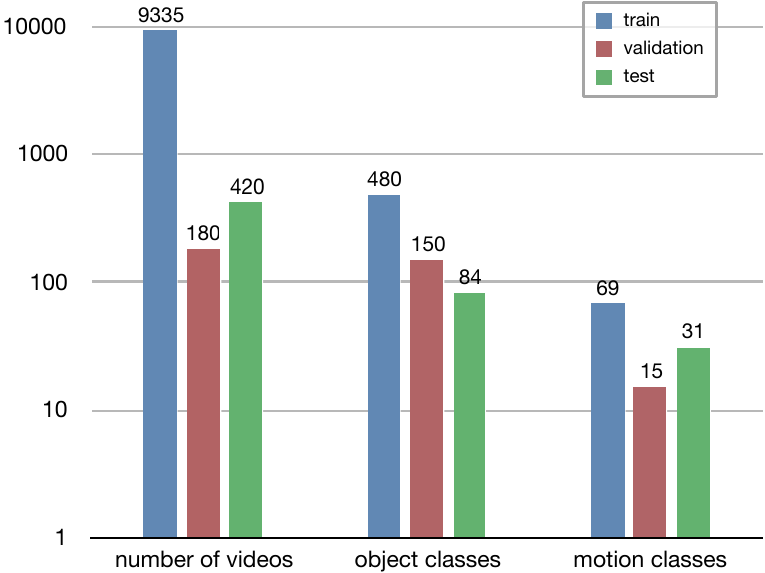}
  \caption{Dataset splits of GOT-10k. Except for the \textit{person} class, all object classes between training and test videos are not overlapped; while for \textit{person}s, the motion classes between training and test are not overlapped. The validation set is randomly sampled from training videos with uniform distribution across different object classes.}
  \label{fig:dataset_splits}
\end{figure}

We split the GOT-10k dataset into unified training, validation and test sets to enable fair comparison of tracking approaches. Unlike many other machine learning applications~\cite{imagenet2009,youtubebb2017}, the splitting of generic object tracking dataset is not straightforward (i.e., by randomly sampling a proportion of data). For one thing, we expect the evaluation results to reflect the generalization ability of different approaches across a wide range of objects and scenarios. To achieve this, an explicit domain gap must be established between training and test videos. For another, we do not need thousands of videos to assess a tracking algorithm. Furthermore, the evaluation of trackers is very time-consuming, thus it is advantageous to keep the test set relatively compact~\cite{vot2013,vot2018}.

With the first consideration, we introduce the \textit{one-shot protocol} and set up a strict rule that the object classes in training and test videos are \textit{non-overlapped}; the \textit{person} class, however, is treated as an exception. Unlike other classes of objects, \textit{person}s embrace a rich collection of motion patterns like \textit{jogging, swimming, skiing, crawling, cycling, diving, equitation, judo,} and \textit{surfing}, to name a few. Each motion pattern represents a special combination of challenges and thus form a problem domain. We argue that an object class with such a variety is of great interest in both training and evaluation, therefore we include \textit{person}s in both training and test sets. To also introduce a domain gap, we ensure in the splitting that the motion classes of \textit{person}s between training and test sets are \textit{non-overlapped}.

To address the second consideration, we carry out a series of experiments to find a reliable while compact test set. We take the test set as a random variable and draw its samples from a large pool of around a thousand videos. Then we run tracking experiments on each sample and evaluate the performance. We run 25 trackers on each sample and rank them according to their AO (average overlap) scores. The ranking stability, i.e. the standard deviation of methods' ranks is used as the indicator for evaluation stability. Results are visualized in Figure~\ref{fig:stdv}. We analyze each influence factor in the following.

\noindent\textbf{Impact of video number.} We vary the number of test videos from 5 to 1000, with a step of 5. Figure~\ref{fig:stdv_video_num} shows that the standard deviation of ranks significantly decreases as video number increases, which indicates an improved evaluation stability. Considering the stability-efficiency trade-offs, we set the video number to 420 in our benchmark, where the average standard deviation of method ranks is below 1.0 and a fair evaluation stability can be achieved. Although further increasing the video number (e.g., 1000 videos) can further improve our ranking stability, the improvement is however marginal (less than 0.5 reduction in the standard deviation of ranks) at a much higher evaluation cost.

\noindent\textbf{Impact of object classes.} We fix the video number to 420 and change the sampled object classes from 5 to 115, results are visualized in Figure~\ref{fig:stdv_obj_class}. We observe an obvious downward trend of the standard deviation of methods' ranks as more object classes are included, verifying the importance of test set variety on the leaderboard stability. We include 84 object classes in our test data (which are not overlapped with our training classes), where the average standard deviation is around $1.1$ in Figure~\ref{fig:stdv_obj_class} at such a setting.

\noindent\textbf{Impact of motion classes.} With the number of videos fixed at 420, we vary the number of motion classes from 5 to 40. Figure~\ref{fig:stdv_mot_class} shows the impact of motion classes on the ranking stability. The stability generally improves as more classes of motion are included in the test set. We include 31 motion classes in our test set, where the standard deviation of methods' ranks is around $1.1$ in Figure~\ref{fig:stdv_mot_class}.

\noindent\textbf{Impact of repetition time.} Many tracking benchmarks require trackers to run multiple times (e.g., 15 times in VOT challenges~\cite{vot2017}) on their datasets to ensure a reliable evaluation. This significantly multiplies the evaluation costs. We quantitatively assess the impact of repetition time on the ranking stability here. We run trackers multiple times and plot the relationship between repetition times and the standard deviation of methods' ranks in Figure~\ref{fig:stdv_repetition}. The video number is fixed to 420 and the repetition time is increased from 1 to 30. We find the contribution of increasing repetition time to the evaluation stability is negligible (at the order of 0.1) when evaluated on our large test set. Considering the stochastic character of many trackers, we set the number of repetitions to 3, which is fairly enough for a stable evaluation.

According to the above analysis, the final splits of GOT-10k dataset are summarized in Figure~\ref{fig:dataset_splits}. The test set contains 420 videos, 84 classes of moving objects and 31 forms of motion, where a reasonably stable ranking can be observed at such a setting (Figure~\ref{fig:stdv}). Except for the \textit{person} class, all object classes between training and test videos are \textit{non-overlapped}; while for \textit{person}s, the motion classes between training and test sets are not overlapped.
We limit the maximum number of videos per class to 8 (which accounts for only 1.9\% of the test set size) to avoid larger-scale classes dominant the evaluation results.
The validation set is selected by randomly sampling 180 videos from the training subset, with a uniform probability across different object classes. For each stochastic tracker, we run 3 times of experiments and average the scores to ensure a reproducible evaluation.

\section{Experiments}\label{sec:experiments}

We carry out extensive experiments on GOT-10k and analyze the results in this section. We expect the baseline performance to offer an aspect of overall difficulty of GOT-10k as well as a point of comparison for future works. We also discuss the challenges of real-world tracking and the impact of training data on the performance of deep trackers.

\subsection{Baseline Models}\label{sec:baseline_models}

We consider recent deep learning and correlation filters based trackers in this work, since they prevail in recent tracking benchmarks and challenges~\cite{otb2015,vot2017,vot2018,datasetvotlongterm2018}. We also evaluate some traditional pioneer works for completeness. The baseline trackers assessed in our benchmark are briefly described in the following.

\noindent\textbf{Deep learning based trackers.}\label{sec:deep_baselines}
Deep learning has boosted the tracking performance largely in recent years~\cite{policy2011,siamesefc2016,mdnet2016,siamrpn2018}. In this work, we consider siamese tracker SiamFC~\cite{siamesefc2016} and its variants CFNet~\cite{cfnet2017}, DCFNet~\cite{dcfnet2017}, DSiam~\cite{dsiam2017}, SASiam~\cite{sasiam2018}, and SiamDW~\cite{siamdw2019}; recurrent tracker MemTracker~\cite{memtracker2018}; meta-learning based tracker MetaTracker~\cite{metatracker2018}; attentive tracker DAT~\cite{dat2018}; and other convolutional neural networks based approaches MDNet~\cite{mdnet2016}, GOTURN~\cite{goturn2016}, CF2~\cite{cf22015}, and RT-MDNet~\cite{rtmdnet2018}.
To ensure a fair and meaningful comparison, we retrain each of these algorithms on GOT-10k's training set with \textbf{no extra training data used}. Nevertheless, as widely practiced in many computer vision tasks~\cite{fasterrcnn2015}, we allow the methods to use ImageNet~\cite{imagenet2009} pretrained weights for model initialization before training on GOT-10k. We adopt default options of these deep trackers, where the backbones of DSiam, SASiam, SiamDW, MetaTracker, MDNet, GOTURN, CF2, and RT-MDNet are fully or partially initialized from ImageNet pretrained weights, while all other trackers are trained from scratch.

\noindent\textbf{Correlation filters based trackers.}
We consider pioneer works on correlation filters based tracking: CSK~\cite{csk2012}, KCF~\cite{kcf_tpami2015} and their variants ColorCSK~\cite{dat2015}, SAMF~\cite{samf2014}, DSST~\cite{dsst2014}, Staple~\cite{staple2016}, SRDCF~\cite{srdcf2015}, SRDCFdecon~\cite{srdcfdecon2016}, CCOT~\cite{ccot2016}, BACF~\cite{bacf2017}, STRCF~\cite{strcf2018}, DeepSTRCF~\cite{strcf2018}, ECO~\cite{eco2017}, and LDES~\cite{ldes2019} in our evaluation.
MOSSE~\cite{mosse2010} is considered the first approach to introduce correlation filters to object tracking. CSK introduces non-linear kernels to correlation filters. ColorCSK, KCF and Staple extend CSK with multi-channel visual features. SAMF and DSST propose efficient scale searching schemes for correlation filters tracking. To tackle with boundary effects, SRDCF and SRDCFdecon apply spatial regularization on learned filters, while BACF uses center cropping on larger shifted samples to remove the impact of boundaries. STRCF introduces temporal regularization to SRDCF to handle large appearance variations. CCOT presents a continuous convolution operator to integrate multi-layer features of convolutional neural networks, while ECO raises both the speed and accuracy of CCOT with several improvements. LDES extends correlation filters with the ability to estimate scale and rotation variations.

\noindent\textbf{Traditional trackers.}
In addition to popular deep learning and correlation filters based methods, we also evaluate some traditional pioneer works. They include generative methods LK~\cite{lk2002}, IVT~\cite{ivt2008}, and L1APG~\cite{l1apg2012} and discriminative method MEEM~\cite{meem2014}. Although these trackers are not state-of-the-art in recent benchmarks, their algorithm designs may inspire future works, thus we also include them in our benchmark for completeness.

For all the evaluated models, we use their public code with default parameter settings throughout our experiments. Although adjusting parameters on our validation set may improve their performance, it requires a huge amount of work. In this respect, the evaluation results in this work can be viewed as a lower bound of these algorithms.
\begin{table*}
  \caption{Overall tracking results of baseline trackers on GOT-10k. The trackers are ranked by their mean average overlap (mAO) scores. The first-, second- and third-place trackers are labeled with red, blue and green colors respectively. The \textit{Properties} column denotes the attributes of different trackers that are split into: correlation filters (yes/no), deep learning (yes/no) and feature representation (CNN - Convolutional Neural Networks, HOG - Histogram of Gradients, CN - Color Names, CH - Color Histogram, Raw - Raw intensity, IIF - Illumination Invariant Features).}
  \label{tab:overall_performance}
  \footnotesize
  \begin{center}
    \begin{tabular}{|
      >{\raggedright\arraybackslash} m{3cm} |
      >{\centering\arraybackslash} m{1cm} |
      >{\centering\arraybackslash} m{1cm} |
      >{\centering\arraybackslash} m{1cm} |
      >{\raggedleft\arraybackslash} m{1.5cm} |
      >{\centering\arraybackslash} m{1cm} |
      >{\centering\arraybackslash} m{1cm} |
      >{\centering\arraybackslash} m{2.5cm} |
      >{\centering\arraybackslash} m{1.5cm} |
    }
    \hline
    \multirow{2}*{\textbf{Tracker}} & \multicolumn{4}{c|}{\textbf{Performance}} & \multicolumn{3}{c|}{\textbf{Properties}} & \multirow{2}*{\textbf{Venue}}\\
    \cline{2-8} & \textbf{mAO} & \textbf{$\text{mSR}_{50}$} & \textbf{$\text{mSR}_{75}$} & \textbf{Speed (fps)} & \textbf{CF} & \textbf{DL} & \textbf{Repr.} &\\
    \hline
    \hline
    1. MemTracker~\cite{memtracker2018} & \textbf{\textcolor{red}{0.460}} & \textbf{\textcolor{red}{0.524}} & \textbf{\textcolor{red}{0.193}} & 0.353@GPU & & \checkmark & CNN & ECCV'18\\
    2. DeepSTRCF~\cite{strcf2018} & \textbf{\textcolor{blue}{0.449}} & 0.481 & \textbf{\textcolor{green}{0.169}} & 1.07@GPU & \checkmark & \checkmark & CNN, HOG, CN & CVPR'18\\
    3. SASiamP~\cite{sasiam2018} & \textbf{\textcolor{green}{0.445}} & \textbf{\textcolor{green}{0.491}} & 0.165 & 25.4@GPU & & \checkmark & CNN & CVPR'18\\
    4. SASiamR~\cite{sasiam2018} & 0.443 & \textbf{\textcolor{blue}{0.492}} & 0.160 & 5.13@GPU & & \checkmark & CNN & CVPR'18\\
    5. SiamFCv2~\cite{cfnet2017} & 0.434 & 0.481 & \textbf{\textcolor{blue}{0.190}} & 19.6@GPU & & \checkmark & CNN & CVPR'17\\
    6. GOTURN~\cite{goturn2016} & 0.418 & 0.475 & 0.163 & \textbf{\textcolor{red}{70.1@GPU}} & & \checkmark & CNN & ECCV'16\\
    7. DSiam~\cite{dsiam2017} & 0.417 & 0.461 & 0.149 & 3.78@GPU & & \checkmark & CNN & ICCV'17\\
    8. SiamFCIncep22~\cite{siamdw2019} & 0.411 & 0.456 & 0.154 & 12@GPU & & \checkmark & CNN & CVPR'19\\
    9. DAT~\cite{dat2018} & 0.411 & 0.432 & 0.145 & 0.0774@GPU & & \checkmark & CNN & NIPS'18\\
    10. CCOT~\cite{ccot2016} & 0.406 & 0.415 & 0.161 & 0.57@CPU & \checkmark & & CNN & ECCV'16\\
    11. MetaSDNet~\cite{metatracker2018} & 0.404 & 0.423 & 0.156 & 0.526@GPU & & \checkmark & CNN & ECCV'18\\
    12. RT-MDNet~\cite{rtmdnet2018} & 0.404 & 0.424 & 0.147 & 7.85@GPU & & \checkmark & CNN & ECCV'18\\
    13. SiamFCNext22~\cite{siamdw2019} & 0.398 & 0.430 & 0.151 & 12.2@GPU & & \checkmark & CNN & CVPR'19\\
    14. ECO~\cite{eco2017} & 0.395 & 0.407 & 0.170 & 2.21@CPU & \checkmark & & CNN, HOG & CVPR'17\\
    15. SiamFC~\cite{siamesefc2016} & 0.392 & 0.426 & 0.135 & \textbf{\textcolor{blue}{32.6@GPU}} & & \checkmark & CNN & ECCV'16\\
    16. SiamFCRes22~\cite{siamdw2019} & 0.385 & 0.401 & 0.143 & 14.5@GPU & & \checkmark & CNN & CVPR'19\\
    17. CF2~\cite{cf22015} & 0.379 & 0.380 & 0.134 & 0.865@GPU & & \checkmark & CNN & ICCV'15\\
    18. STRCF~\cite{strcf2018} & 0.377 & 0.387 & 0.151 & 3.06@CPU & \checkmark & & HOG, CN & CVPR'18\\
    19. DCFNet~\cite{dcfnet2017} & 0.364 & 0.378 & 0.144 & 18.9@GPU & \checkmark & \checkmark & CNN & ArXiv'17\\
    20. CFNetc2~\cite{cfnet2017} & 0.364 & 0.365 & 0.150 & 30.3@GPU & \checkmark & \checkmark & CNN & CVPR'17\\
    21. ECOhc~\cite{eco2017} & 0.363 & 0.359 & 0.154 & \textbf{\textcolor{green}{34.7@CPU}} & \checkmark & & HOG, CN & CVPR'17\\
    22. LDES~\cite{ldes2019} & 0.359 & 0.368 & 0.153 & 1.23@CPU & \checkmark & & HOG, CH & AAAI'19\\
    23. MDNet~\cite{mdnet2016} & 0.352 & 0.367 & 0.137 & 0.951@GPU & & \checkmark & CNN & CVPR'16\\
    24. BACF~\cite{bacf2017} & 0.346 & 0.361 & 0.149 & 3.22@CPU & \checkmark & & HOG & ICCV'17\\
    25. CFNetc1~\cite{cfnet2017} & 0.343 & 0.341 & 0.136 & \textbf{\textcolor{green}{32.6@GPU}} & \checkmark & \checkmark & CNN & CVPR'17\\
    26. CFNetc5~\cite{cfnet2017} & 0.337 & 0.321 & 0.126 & 21.6@GPU & \checkmark & \checkmark & CNN & CVPR'17\\
    27. Staple~\cite{staple2016} & 0.332 & 0.333 & 0.135 & 6.91@CPU & \checkmark & & HOG, CN & CVPR'16\\
    28. SAMF~\cite{samf2014} & 0.330 & 0.331 & 0.130 & 1.47@CPU & \checkmark & & HOG, CN, Raw & ECCVW'14\\
    29. MEEM~\cite{meem2014} & 0.320 & 0.305 & 0.093 & 11.1@CPU & & & Lab, IIF & ECCV'14\\
    30. DSST~\cite{dsst2014} & 0.317 & 0.314 & 0.136 & 1.03@CPU & \checkmark & & HOG & BMVC'14\\
    31. SRDCF~\cite{srdcf2015} & 0.312 & 0.310 & 0.134 & 2.26@CPU & \checkmark & & HOG & ICCV'15\\
    32. SRDCFdecon~\cite{srdcfdecon2016} & 0.310 & 0.311 & 0.139 & 1.59@CPU & \checkmark & & HOG & CVPR'16\\
    33. fDSST~\cite{fdsst_tpami2017} & 0.289 & 0.278 & 0.121 & 3.39@CPU & \checkmark & & HOG & TPAMI'17\\
    34. KCF~\cite{kcf_tpami2015} & 0.279 & 0.263 & 0.099 & 11.5@CPU & \checkmark & & HOG & TPAMI'15\\
    35. ColorCSK~\cite{dat2015} & 0.272 & 0.249 & 0.092 & 4.52@CPU & \checkmark & & CH & CVPR'15\\
    36. CSK~\cite{csk2012} & 0.264 & 0.236 & 0.085 & \textbf{\textcolor{blue}{40.5@CPU}} & \checkmark & & Raw & ECCV'12\\
    37. L1APG~\cite{l1apg2012} & 0.259 & 0.257 & 0.108 & 12.6@CPU & & & Raw & CVPR'12\\
    38. LK~\cite{lk2002} & 0.237 & 0.202 & 0.068 & 1.44@CPU & & & Raw & CVPR'02\\
    39. IVT~\cite{ivt2008} & 0.171 & 0.114 & 0.031 & \textbf{\textcolor{red}{47.3@CPU}} & & & Raw & IJCV'08\\
    \hline
    \end{tabular}
  \end{center}
\end{table*}
\begin{figure*}[!ht]
  \centering
  \begin{subfigure}[b]{0.45\textwidth}
      \includegraphics[width=\textwidth]{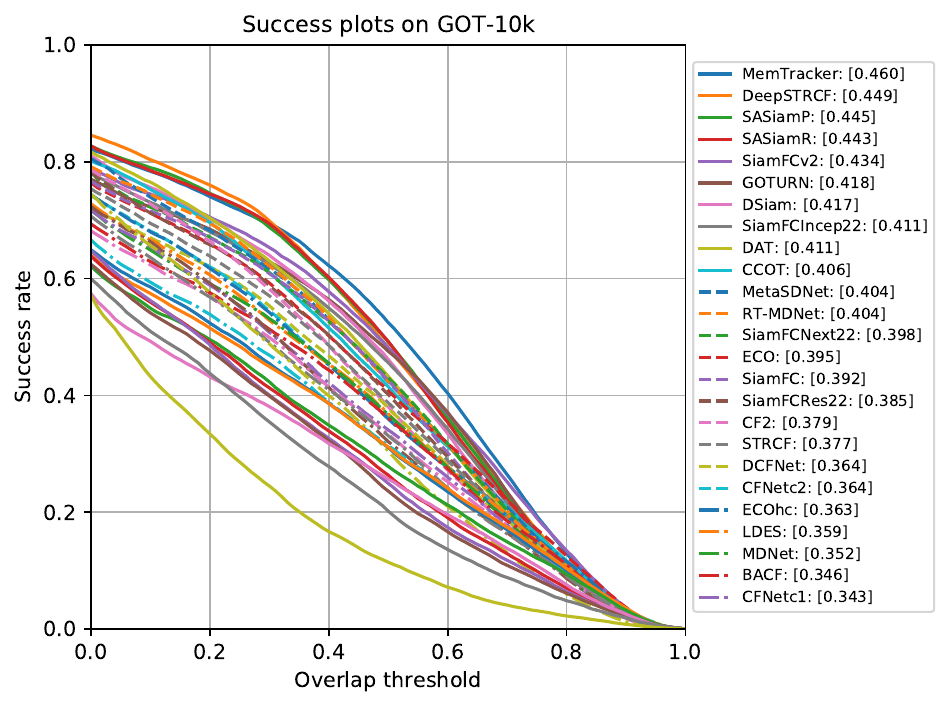}
      \caption{Success plots on GOT-10k.}
      \label{fig:overalll_performance_got10k}
  \end{subfigure}
  \begin{subfigure}[b]{0.45\textwidth}
      \includegraphics[width=\textwidth]{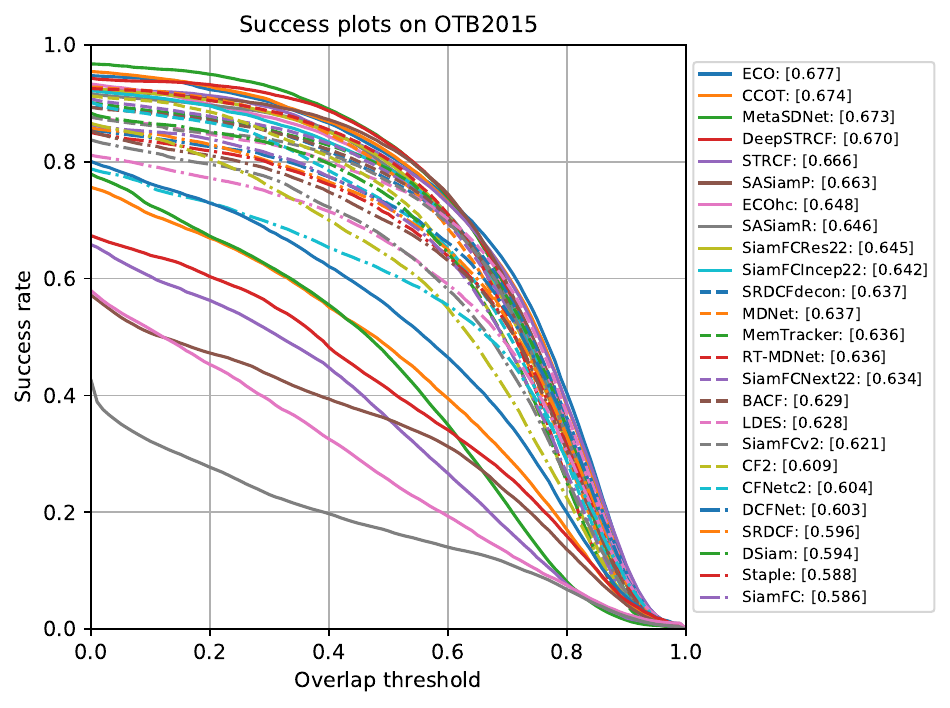}
      \caption{Success plots on OTB2015.}
      \label{fig:overalll_performance_otb}
  \end{subfigure}
  \caption{Success curves of 39 baseline trackers on GOT-10k (420 sequences) and OTB2015 (100 sequences) benchmarks, ranked by their mean average overlap (mAO) scores. For clarity, only the top 25 trackers are shown.}
  \label{fig:overall_performance}
\end{figure*}
\subsection{Evaluation Methodology}\label{sec:evaluation_methodology}
In this work, we prefer to employ simple metrics with clear meaning for the evaluation of trackers. We choose the widely used average overlap (AO) and success rate (SR) as our indicators. The AO denotes the average of overlaps between all groundtruth and estimated bounding boxes, while the SR measures the percentage of successfully tracked frames where the overlaps exceed a threshold (e.g., 0.5). The AO is recently proved~\cite{votmetrics2014,votmetrics2016,otb2015} to be equivalent to the area under curve (AUC) metric employed in OTB~\cite{otb2013,otb2015}, NfS~\cite{datasetvotspeed2017}, UAV~\cite{datasetuav2016}, TrackingNet~\cite{datasettrackingnet2018}, and LaSOT~\cite{datasetlasot2018} datasets. Besides, the expected average overlap (EAO) metric used for overall ranking in VOT challenges is an approximation of AO on larger video pool. The SR metric is also used in the OTB2015~\cite{otb2015} and OxUvA~\cite{datasetvotlongterm2018} datasets. It clearly indicates how many frames are tracked or lost under certain precision, which is the concern of many applications.

Unlike existing tracking benchmarks that directly average over sequence-wise scores to obtain the final performance, which totally ignores the potential class-imbalance in evaluation (i.e., dominant classes with more sequences result in higher weights), we propose class-balanced metrics. Taking AO as an example, the class-balanced metric mAO (mean average overlap) is calculated as:
\begin{eqnarray}
  \text{mAO} = \frac{1}{C} \sum_{c=1}^C (\frac{1}{|S_c|} \sum_{i\in S_c} \text{AO}_{i}).
\end{eqnarray}
where $C$ is the class number, $S_c$ denotes the subset of sequences belonging to the $c$th class, while $|S_c|$ is the subset scale. In the formula, class-wise AOs are computed first (in the bracket) and then averaged to obtain the final score, with equal treatment to different classes. The same principle is applied to SR, where the mSR is computed by averaging the SRs over different classes. We use two overlap thresholds 0.5 and more strict 0.75 for calculating the mSR.

The success curve~\cite{otb2013,otb2015} is used in our benchmark to visualize the tracking results. Each point of the success curve shows the percentage of frames where the overlaps exceed a threshold. The success curve offers a continuous measurement of tracking results ranging from robustness (lower overlap rate but more tracked frames) to accuracy (higher overlap rate)~\cite{otb2015,unveiling2018}. As discussed in Section~\ref{sec:dataset_splitting}, for each stochastic method, we run 3 times of tracking experiments and average the evaluation results to achieve a stable evaluation.
\begin{figure*}
  \centering
  \includegraphics[width=180mm]{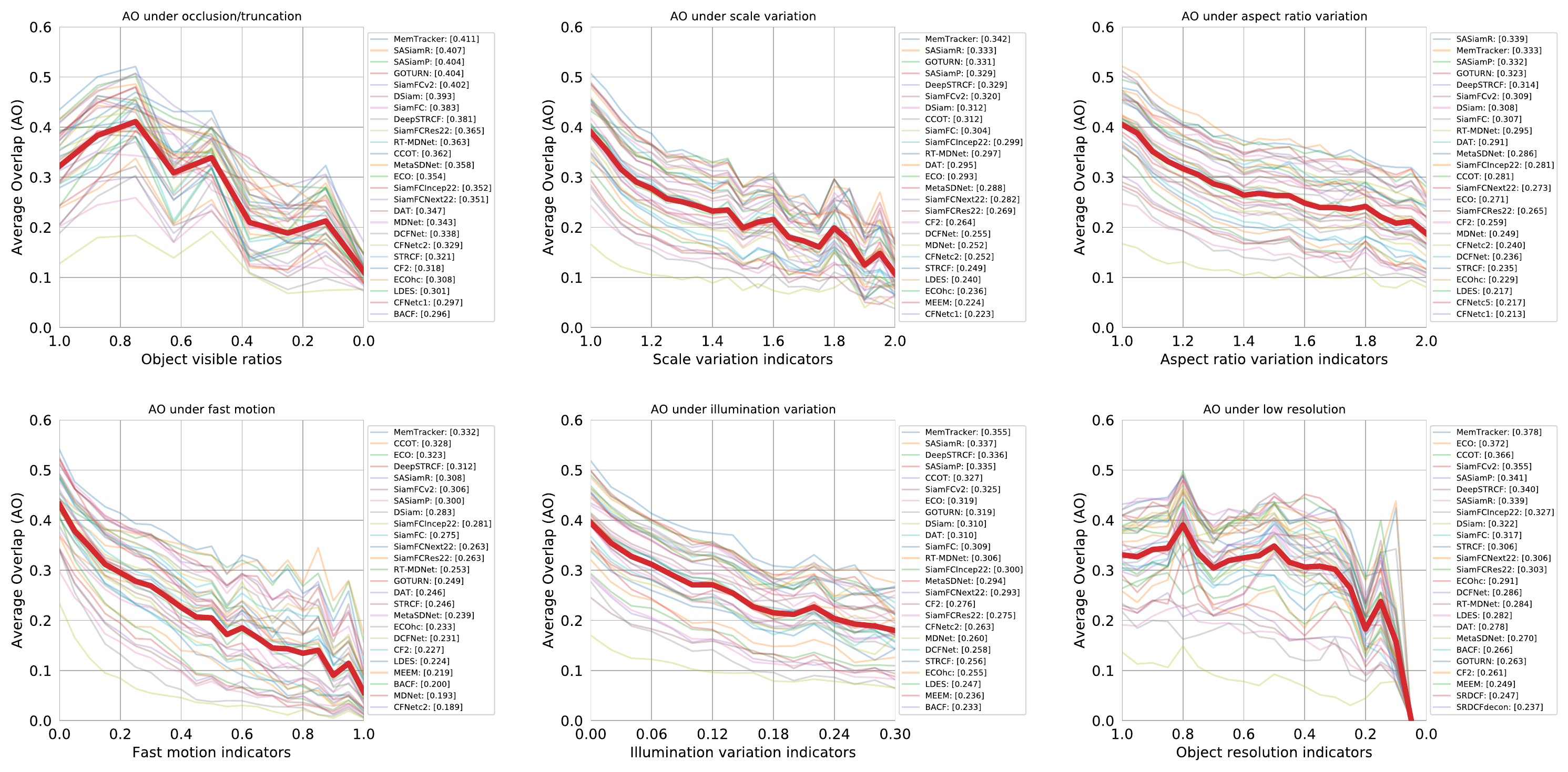}
  \caption{Performance of baseline trackers under 6 challenging attributes: \textit{occlusion/truncation, scale variation, aspect ratio variation, fast motion, illumination variation} and \textit{low resolution targets}. Each attribute is quantified as a continuous difficulty indicator, which can be directly computed from the annotations, as introduced in Section~\ref{sec:evaluation_by_challenges}. The thick red curves represent the average performance over trackers. The final AO score of each tracker at the legend is computed by averaging the results on the top 20\% hardest frames. For clarity, only the top 25 trackers are shown in the plots.}
  \label{fig:challenge_curves}
\end{figure*}
\subsection{Overall Performance}
We employ the mean average overlap (mAO) and mean success rate (mSR) for the overall evaluation of trackers, as described in Section~\ref{sec:evaluation_methodology}. For deep trackers, we retrain each of them on GOT-10k to achieve a fair comparison (see Section~\ref{sec:deep_baselines} for details). All experiments are run on a server with a 56 core Intel(R) Xeon(R) 2.0GHz CPU and 4 GeForce GTX TITAN X graphic cards. Table~\ref{tab:overall_performance} illustrates the evaluation results of all baseline models, ranked by their mAO scores. Figure~\ref{fig:overalll_performance_got10k} shows their success curves. For comparison, we also re-evaluate the baseline methods on OTB2015 with the same weights retrained on GOT-10k, results are shown in Figure~\ref{fig:overalll_performance_otb}.

The top 5 trackers on GOT-10k are MemTracker, DeepSTRCF, SASiamP, SASiamR, and SiamFCv2. Except for DeepSTRCF, which is a correlation filters based method, all other trackers are built upon fully convolutional siamese networks~\cite{siamesefc2016}. MemTracker achieves the best performance and it outperforms the 2nd placed trackers by 1.1\%, 3.3\% and 0.3\% in terms of mAO, $\text{mSR}_{50}$ and $\text{mSR}_{75}$, respectively, while the other 4 trackers achieve close performance. The following 5 trackers are GOTURN, DSiam, SiamFCIncep22, DAT, and CCOT, where DSiam and SiamFCIncep22 are siamese networks based methods, GOTURN is a siamese bounding box regression tracker, DAT is built upon the attention mechanism, while CCOT is based on correlation filters and it uses pretrained CNNs for feature extraction. Among traditional trackers using only hand-crafted features, STRCF, ECOhc, LDES, BACF, and Staple obtain the top five evaluation scores. Although no deep features used, their results are comparable or even better than some deep trackers, such as MDNet and CFNet. From the evaluation results, we observe that the highest mAO score on GOT-10k only reaches 46.0\%, suggesting that tracking in real-world unconstrained videos is difficult and still not solved.

By comparing the evaluation results on GOT-10k and OTB2015, we observe a significant change in the order of methods. For example, ECO achieves the best performance on OTB2015 but it performs much worse on GOT-10k. On the other hand, GOTURN obtains very low scores on OTB2015 while it outperforms most trackers on GOT-10k. In addition, by comparing some methods with their improved versions, we can also observe the differences between OTB2015 and GOT-10k evaluation results. For example, ECO improves CCOT in several ways and it outperforms CCOT on OTB2015, but on GOT-10k its mAO score is lower than CCOT. Similar phenomenon can also be observed by comparing DSST with its improved version fDSST, and SRDCF with its improved method SRDCFdecon. One possible reason for such differences may be that some high-performance trackers are overfitted to small datasets, or they need a certain amount of hyperparameter tuning to achieve better performance, while methods with straightforward frameworks may have better generalization ability in challenging scenarios.

The \textit{Speed (fps)} column in Table~\ref{tab:overall_performance} shows the tracking speeds of different approaches. Among the GPU trackers, GOTURN achieves the highest speed of 70.1 frames per second (fps), followed by SiamFC (32.6 fps) and CFNetc1 (32.6 fps). GOTURN and SiamFC benefit from their extremely simple architectures and tracking pipelines, while CFNet reformulates the efficient correlation filters in an end-to-end learnable module to achieve high-speed tracking. On the CPU platform, IVT is the fastest tracker that runs at around 47.3 fps, followed by CSK and ECOhc. The efficiency of the training and inference of correlation filters plays a key role in the high speed of CSK and ECOhc, while the fast incremental subspace updating scheme contributes to the efficiency of IVT.
Note the tracking speeds evaluated on GOT-10k are usually lower than their reported results on OTB and VOT. This is because the resolutions of videos and objects in GOT-10k are much higher ($3\sim 9$ times) than OTB and VOT datasets.
While these high-resolution videos slow down almost all trackers due to more computational costs spent on larger image processing (e.g., cropping, resizing, and padding.), those approaches that downsample the search regions to a fixed size (e.g., most siamese trackers such as SiamFC, SASiamP and CFNet) are less affected; while those approaches with search area sizes proportional to the objects' resolutions (e.g., some correlation filters based approaches such as CSK, KCF and DSST) run significantly slower on our videos.
%

\subsection{Evaluation by Challenges}\label{sec:evaluation_by_challenges}

Although the overall performance indicates the general quality of trackers, it cannot differentiate them according to different attributes and thus reflect the strengths and weaknesses of each method. In this section, we analyze the performance of trackers in terms of different challenges.

Instead of subjectively labeling attributes for each frame or video, as is widely practiced in many existing benchmarks~\cite{otb2013,otb2015,datasetlasot2018}, we prefer a more objective and scalable way for challenge labeling, such that the labels are reproducible and unbiased of the annotators. Similar to ~\cite{datasetvotlongterm2018}, we set up a set of continuous difficulty indicators for each video frame that can be directly computed from the annotations, which are defined as follows:

\noindent\textbf{Occlusion/Truncation.} The occlusion/truncation indicator can be directly deduced from the labeling of visible ratios $v$. We define the degree of occlusion/truncation as $(1 - v)$.

\noindent\textbf{Scale variation.} The scale variation of a target in $i$th frame is measured by $\max\{s_i / s_{i - T},~s_{i - T} / s_i\}$, where $s_i = \sqrt{w_i h_i}$ denotes the object size, while $T$ is a time span for assessing the scale variation over the last $T$ frames. We set $T=5$ in our evaluation.

\noindent\textbf{Aspect ratio variation.} Object deformation and rotation can be characterized by the change of aspect ratios. We measure the degree of aspect ratio variation in frame $i$ as $\max\{r_i / r_{i - T},~r_{i - T} / r_i\}$, where $r_i = h_i / w_i$ and $T$ is a time span fixed to $T=5$.

\noindent\textbf{Fast motion.} We measure the object motion speed at the $i$th frame relative to its size as:
\begin{eqnarray}
  d_i = \frac{\|p_i - p_{i - 1}\|_2}{\sqrt{s_i s_{i - 1}}}
\end{eqnarray}
where $p_i$ denotes the object center location and $s_i = \sqrt{w_i h_i}$ represents the object size.

\noindent\textbf{Illumination variation.} The degree of illumination variation in each frame can be measured by the change of average colors $u_i = \|c_i - c_{i - 1}\|_1$, where $c_i$ is the average object color (with RGB channels normalized to $[0, 1]$) at frame $i$.

\noindent\textbf{Low resolution objects.} Objects with small sizes may affect the tracking performance since less fine-grained features can be extracted from them. We measure the resolution of an object by comparing its size with the median object size in our dataset. Formally, the indicator is defined as $s_i / s^{\text{median}}$, where $s_i = \sqrt{w_i h_i}$ and $s^{\text{median}}$ is the median of object sizes over all frames in our test data. We only consider frames with $s_i \leq s^{\text{median}}$ in the evaluation.

When all frame-wise difficulty indicators are acquired, we divide their values into several discrete intervals and compute the AO scores on the frame subsets within each interval. The performance w.r.t. increased difficulty for each challenging attribute is visualized in Figure~\ref{fig:challenge_curves}.
From the figure, we observe a discernible decline in tracking performance for attributes \textit{fast motion, aspect ratio variation, scale variation} and \textit{illumination variation} when difficulty increases. This indicates that tracking under fast object state (position, scale, and orientation) and appearance (pose and illumination conditions) changes are still challenging for current trackers.
We also find that the tracking performance decreases rapidly when objects' visible ratios become lower (i.e., below 0.5), suggesting the difficulty in robust tracking when targets' features are truncated or influenced by external distractors.
For attribute \textit{object resolution}, we observe that the AO scores do not change much when objects' resolutions are relatively high (i.e., above 0.3); but drop rapidly when their resolutions become very low (i.e., below 0.3). This indicates the difficulty of tracking low resolution or very small moving objects.
In general, for all 6 challenging attributes, the increase of difficulty causes significant performance degradation for almost all baseline trackers.

Among the baseline trackers, the performance of GOTURN improves on \textit{scale variation} and \textit{aspect ratio variation} subsets, where its ranking increases by 3 and 2 places, respectively, compared to its ranking on the full test set. The improvement can be attributed to its ability of regressing to arbitrarily sized bounding boxes. ECO and CCOT perform much better than other trackers on \textit{fast motion} and \textit{low resolution} subsets, which is consistent with their designs of large search area (4.5 times the target size) and the aggregation of multi-resolution features (HOG, shallow and deep layer outputs of CNNs). Siamese trackers (e.g., MemTracker, SASiamR, SASiamP, SiamFCv2, DSiam, and SiamFC) perform well on the \textit{occlusion/truncation} subset. This may because that the template matching mechanism is less prone to overfitting to the occluders compared to classifier learning based methods (e.g., DeepSTRCF, CCOT, ECO, etc.). Besides, we notice that MemTracker performs very stable over all 6 challenging attributes. This may be due to its memory mechanism, which is learned offline to find a proper balance between adaptability and overfitting.

\begin{figure*}
  \centering
  \includegraphics[width=180mm]{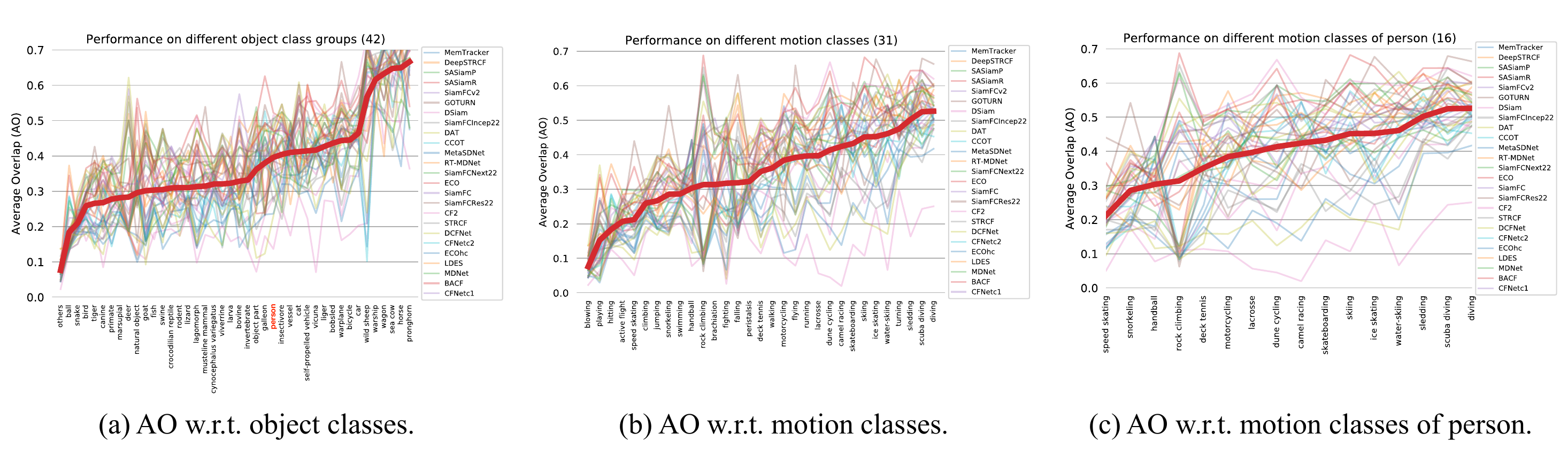}
  \caption{Class-wise performance of baseline trackers on GOT-10k. We divide the test videos of GOT-10k into several subsets according to the object or motion classes and evaluate AO scores on each subset. The thick red curves represent the average performance over baseline trackers. (a) AO w.r.t. object classes. (b) AO w.r.t. motion classes. (c) AO w.r.t. motion classes of \textit{person}. Note for the \textit{person} class, motion classes in training and test sets are not overlapped. Best viewed with zooming in.}
  \label{fig:class_performance}
\end{figure*}

\subsection{Evaluation by Object and Motion Classes}\label{sec:evaluation_by_object_and_motion_classes}

In this section, we discuss the performance of baseline trackers with respect to different object and motion classes. An object or motion class can roughly represent a natural distribution of tracking challenges, thus the analysis of class-wise performance may lead us to find the strengths and bottlenecks of current trackers.

\noindent\textbf{Object classes.} According to the correspondence between our object classes and higher-level class groups (as detailed in Section~\ref{sec:collection}), we divide the 84 object classes in our test set into 42 groups, and evaluate the AO scores of all baseline trackers on each class group. The results are shown in Figure~\ref{fig:class_performance} (a), where the classes are sorted by the average AO scores evaluated on the corresponding subsets.
Although the performance of baseline algorithms varies widely among different object classes, we can still see the differences in overall difficulty between classes.
Generally, the small (e.g., \textit{ball, bird} and \textit{larva}), thin (e.g., \textit{snake, fish} and \textit{lizard}) and fast-moving (e.g., \textit{bird, canine, primate} and \textit{rodent}) objects are usually harder to track than large or slow objects (e.g., \textit{pronghorn, sea cow, wagon} and \textit{warship}). In addition, objects with large deformation (e.g., \textit{snake, primate, crocodilian reptile} and \textit{lizard}) usually lead to lower tracking performance, compared to relatively rigid objects (e.g., all kinds of vehicles). The observations are approximately consistent with the challenge analysis in Section~\ref{sec:evaluation_by_challenges}. We also notice that the \textit{person} class only represents the objects of moderate difficulty, which ranks as the $17$th easiest class out of 42 class groups.

\noindent\textbf{Motion classes.} We evaluate the baseline trackers on subsets of test videos that are labeled with different motion classes. The sorted results are visualized in Figure~\ref{fig:class_performance} (b). From the figure, we observe significant differences in the overall difficulty of different motion classes. In general, fast and dramatic movements (e.g., \textit{blowing, playing, hitting, speed skating} and \textit{jumping}) are usually harder to track than gentle movements (e.g., \textit{diving, scuba diving} and \textit{turning}). Another influencing factor may be related to the potential background clutters. For some of the "easier" motion classes, such as \textit{diving, scuba diving, sledding, ice-skating} and \textit{skiing}, the background may usually be simple with monotonous colors and textures; while for some "harder" motion classes like \textit{blowing, playing, climbing, swimming} and \textit{handball}, the surrounding environment may usually be complex with potential distractors and background clutters, casting further challenges for tracking.

\noindent\textbf{Motion classes of \textit{person}.} The \textit{person} class embrace a rich set of motion forms with each representing a combination of challenges for tracking. It is also the central concern of many industrial applications. Therefore, we separately analyze the tracking performance on \textit{person} in terms of its different motion classes. The results are shown in Figure~\ref{fig:class_performance} (c). Generally, the motion classes with drastic movements (e.g., \textit{speed skating, handball, deck tennis} and \textit{lacrosse}), potential background clutters (e.g., \textit{handball, rock climbing} and \textit{motorcycling}) and potential large deformation (e.g., \textit{rock climbing}) are usually harder to track, where algorithms may drift to distractors or occluders; while the motion classes with potentially cleaner background (e.g., \textit{diving, sledding} and \textit{skiing}) may usually be easier to track.

\begin{figure*}[!ht]
  \centering
  \begin{subfigure}[b]{0.32\textwidth}
      \includegraphics[width=\textwidth]{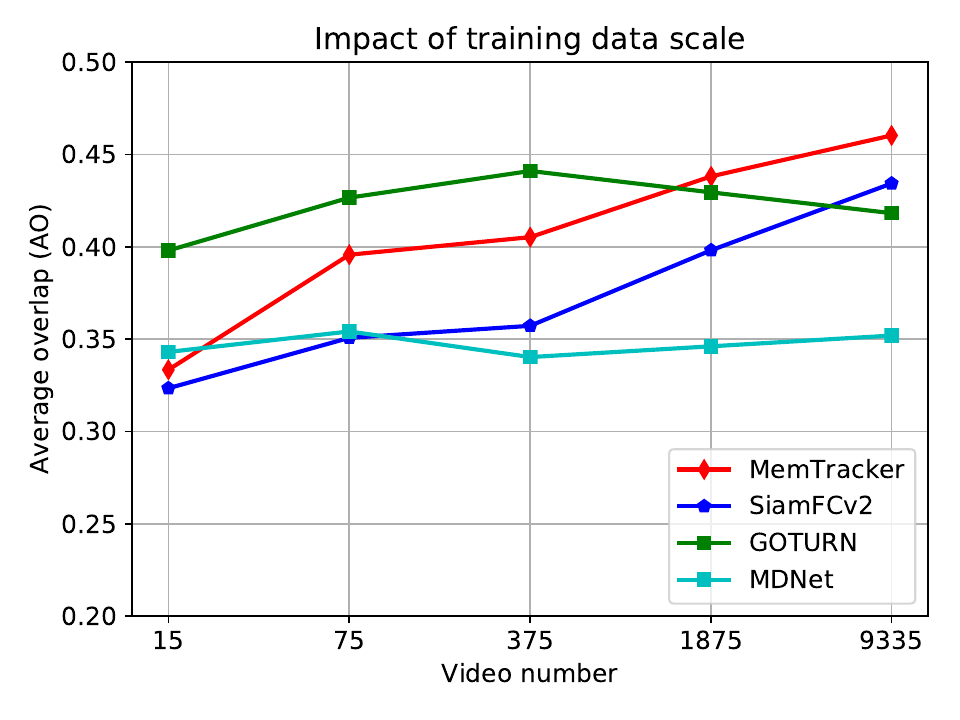}
      \caption{Impact of scale.}
      \label{fig:impact_scale}
  \end{subfigure}
  \begin{subfigure}[b]{0.32\textwidth}
      \includegraphics[width=\textwidth]{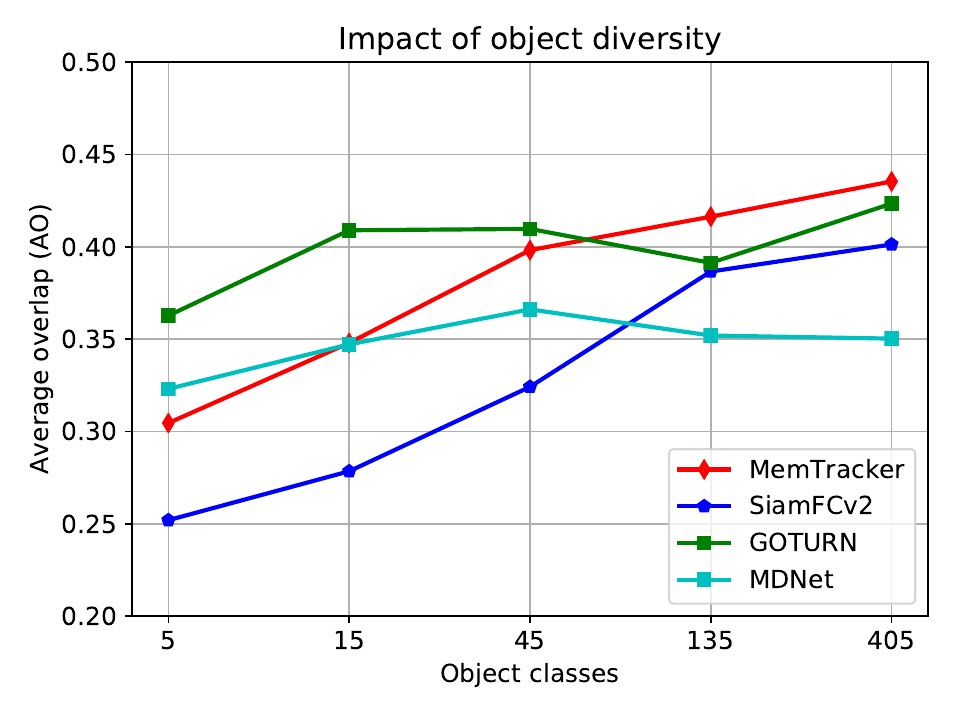}
      \caption{Impact of object diversity.}
      \label{fig:impact_obj_div}
  \end{subfigure}
  \begin{subfigure}[b]{0.32\textwidth}
    \includegraphics[width=\textwidth]{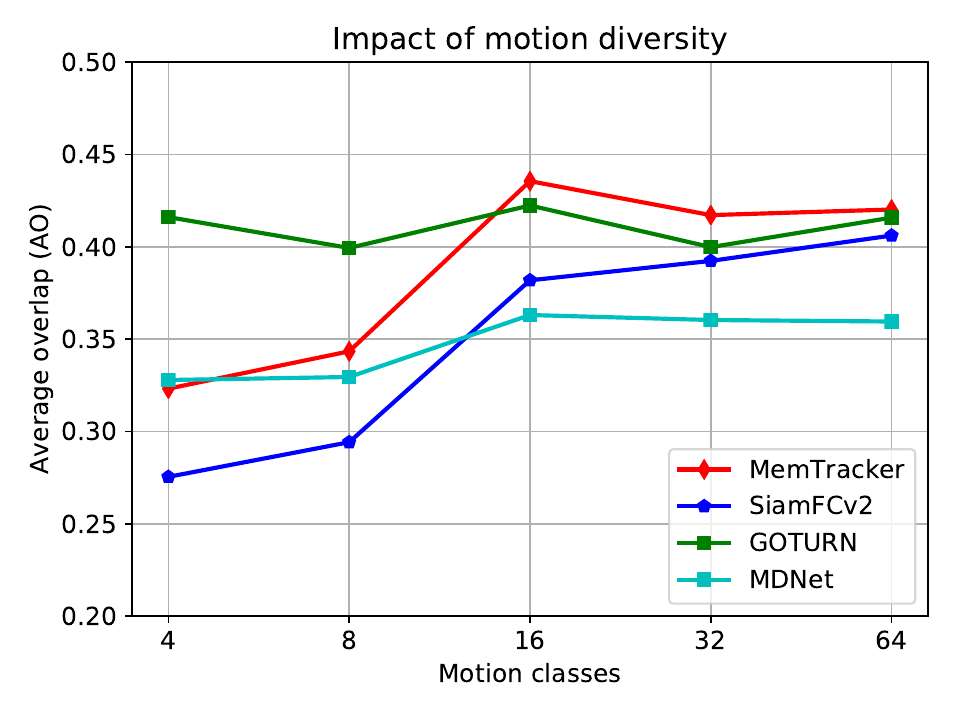}
    \caption{Impact of motion diversity.}
    \label{fig:impact_mot_div}
\end{subfigure}
  \caption{Ablation study on how the scale and diversity of training data impact the performance of deep trackers. (a) Impact of scale. The video number of training data is exponentially increased with a multiplier of 5. (b) Impact of object diversity. The number of object classes is exponentially increased with a multiplier of 3. (c) Impact of motion diversity. The number of motion classes is exponentially increased with a multiplier of 2.}
  \label{fig:impact_training}
\end{figure*}

\subsection{Impact of Training Data}\label{sec:impact_of_training_data}

In this section, we analyze the impact of training data on the performance of deep trackers. Our evaluation covers several aspects including the scale, object diversity, motion diversity, and class distribution of the training data, as well as the class overlaps between training and test sets. We also carry out experiments to assess the cross-dataset (i.e., training on one dataset while testing on another) performance of deep trackers. Unless specified, the ablation study is conducted based on four different deep trackers: MemTracker, SiamFCv2, GOTURN, and MDNet. We discuss the results in the following.

\noindent\textbf{Impact of scale.} We train the four trackers with the video number exponentially increases from 15 to 9335, with a multiplier of 5. The results are shown in Figure~\ref{fig:impact_scale}. We surprisingly find that the dependencies of different deep trackers on data scale differ significantly. The performance of MemTracker and SiamFCv2 significantly improves with the use of more training data, and the trend does not converge at 9335 videos -- it seems that they can benefit from an even larger training set. GOTURN achieves a higher mAO score at the start as more data are used, then the performance converges and even drops as the scale gets larger, which may indicate an under-fitting. By contrast, MDNet seems to be insensitive to the scale of training data -- its performance is roughly saturated at only 15 training videos, while further enlarging data scale only has a minor influence on its evaluation results.
The varying degrees of dependence of these methods on data scale may be partially due to their different trainable model sizes. MemTracker and SiamFCv2 are built upon randomly initialized 5-layer AlexNets~\cite{alexnet2012}, and they allow all parameters to be trainable. Therefore, on small-scale training data, they are prone to over-fitting; while on larger training data, they have more room for improvement. By contrast, MDNet and GOTURN are initialized from ImageNet pretrained weights, and they fix early layers during training. As a consequence, these two methods can deliver good performance with only a small amount of training data, but it may be difficult for them to further benefit from larger training data.

\begin{table}[t]
  \scriptsize
  \begin{center}
      \centering
      \caption{Comparison of the performance of deep trackers on \textit{seen} and \textit{unseen} object classes. The experiments are repeated three times with randomly sampled training and test data. Higher scores are marked in bold.
      }
      \label{tab:seen_unseen}
      \begin{tabular}{|
          >{\raggedright\arraybackslash} m{1.8cm} |
          >{\centering\arraybackslash} m{2.8cm} |
          >{\centering\arraybackslash} m{2.8cm} |}
          \hline
          Trackers & Seen Test Classes & Unseen Test Classes \\
          \hline
          \hline
          MemTracker & \textbf{0.595}/\textbf{0.589}/\textbf{0.612} & 0.588/0.567/0.598 \\
          SiamFCv2 & \textbf{0.584}/\textbf{0.593/}\textbf{0.571} & 0.525/0.569/0.550 \\
          GOTURN & \textbf{0.489}/\textbf{0.491}/\textbf{0.485} & 0.440/0.462/0.457 \\
          MDNet & \textbf{0.443}/0.437/\textbf{0.450} & 0.439/\textbf{0.441}/0.449 \\
          \hline
      \end{tabular}
  \end{center}
\end{table}
\begin{table}[t]
  \scriptsize
  \begin{center}
      \centering
      \caption{Comparison of the tracking performance of deep trackers trained on class-balanced and class-imbalanced data. Higher scores are marked in bold.}
      \label{tab:balanced_imbalanced}
      \begin{tabular}{|
          >{\raggedright\arraybackslash} m{1.6cm} |
          >{\centering\arraybackslash} m{2.9cm} |
          >{\centering\arraybackslash} m{2.9cm} |}
          \hline
          Trackers & Balanced Training Data & Imbalanced Training Data \\
          \hline
          \hline
          MemTracker & \textbf{0.431} & 0.412 \\
          SiamFCv2 & 0.399 & \textbf{0.407} \\
          GOTURN & 0.408 & \textbf{0.421} \\
          MDNet & \textbf{0.357} & 0.340 \\
          \hline
      \end{tabular}
  \end{center}
\end{table}
\begin{table}[t]
  \scriptsize
  \begin{center}
      \centering
      \caption{Comparison of the evaluation stability on balanced and imbalanced test data, assessed by the standard deviation of the ranking of 25 baseline trackers.}
      \label{tab:balanced_imbalanced_test}
      \begin{tabular}{|
          >{\raggedright\arraybackslash} m{0.8cm} |
          >{\centering\arraybackslash} m{2.1cm} |
          >{\centering\arraybackslash} m{2.4cm} |
          >{\centering\arraybackslash} m{1.8cm} |}
          \hline
          Metrics & Balanced Test Data (std. of ranks) & Imbalanced Test Data (std. of ranks) & Abs. Difference between Ranks \\
          \hline
          \hline
          mAO & \textbf{1.102} & 1.213 & 1.259 \\
          mSR & \textbf{1.114} & 1.241 & 1.302 \\
          \hline
      \end{tabular}
  \end{center}
\end{table}
\begin{table}[t]
  \scriptsize
  \begin{center}
      \centering
      \caption{Cross dataset evaluation of deep trackers. The trackers are trained on either ImageNet-VID or GOT-10k datasets and are evaluated on either OTB or GOT-10k datasets. Higher scores are marked in bold.
      }
      \label{tab:cross_dataset}
      \begin{tabular}{|
          >{\raggedright\arraybackslash} m{1.3cm} |
          >{\centering\arraybackslash} m{1.3cm} |
          >{\centering\arraybackslash} m{1.4cm} |
          >{\centering\arraybackslash} m{1.3cm} |
          >{\centering\arraybackslash} m{1.4cm} |}
          \hline
          Trackers & VID $\rightarrow$ OTB & GOT-10k $\rightarrow$ OTB & VID $\rightarrow$ GOT-10k & GOT-10k $\rightarrow$ GOT-10k \\
          \hline
          \hline
          MemTracker & 0.625 & \textbf{0.636} & 0.447 & \textbf{0.460} \\
          SiamFCv2 & 0.613 & \textbf{0.621} & 0.423 & \textbf{0.434} \\
          GOTURN & \textbf{0.427} & 0.413 & 0.396 & \textbf{0.418} \\
          MDNet & \textbf{0.673} & 0.637 & 0.341 & \textbf{0.352} \\
          \hline
      \end{tabular}
  \end{center}
\end{table}

\noindent\textbf{Impact of object diversity.} We fix the number of training videos to 2000 and exponentially vary the number of randomly sampled object classes from 5 to 405. The evaluation results are shown in Figure~\ref{fig:impact_obj_div}.
We observe an upward tendency of the performance of MemTracker, SiamFCv2, and GOTURN as more object classes are introduced in the training data. Noticeably, MemTracker and SiamFCv2 show a steep increase in mAO score (by nearly 15\%) when increasing the number of object classes from 5 to 405, and the trends do not seem to converge at 405 classes, indicating the importance of object diversity in training data on the generalization performance of the two trackers.
In contrast, the performance of MDNet is less affected by the object diversity in the training data.

\noindent\textbf{Impact of motion diversity.} With the training data scale fixed to 2000, we exponentially increase the number of randomly sampled motion classes from 4 to 64. The evaluation results are shown in Figure~\ref{fig:impact_mot_div}. From the figure, we find that the performance of MemTracker and SiamFCv2 significant improves with the increasing number of motion classes. The mAO score of SiamFCv2 reaches a peak value at 16 motion classes, while MemTracker continues to improve until 64 motion classes are introduced. As a comparison, the performance of GOTURN and MDNet is less related to the motion diversity of the training data. This may be because the early layers of GOTURN and MDNet are initialized from ImageNet pre-training weights, which show good generalization capabilities across tasks~\cite{fasterrcnn2015,seg2018,mot_vot2017}, so their tracking performance is less affected by the data diversity. SiamFCv2 and MemTracker are fully trained from scratch and they have a higher dependency on the diversity of training data. Besides, MemTracker learns the modeling of dynamic memories from large video data, which may be the reason that it has a higher dependency on the motion diversity in training data than other trackers.

\noindent\textbf{Training on balanced v.s. imbalanced data.} Like many large-scale datasets~\cite{imagenet2009,youtubebb2017}, the class distribution of GOT-10k is imbalanced. To assess how the class-imbalance impacts the performance of deep trackers, we randomly sample 100 object classes from GOT-10k and construct two training sets: a balanced training set with 20 sequences per class; and an imbalanced training set with the original imbalanced class distribution kept. Both training sets contain 2,000 sequences. We train four deep trackers on these two sets and evaluate their performance on our test data. Results are summarized in Table~\ref{tab:balanced_imbalanced}. From the results, we surprisingly find that the impact of class-imbalance on the tracking performance is not significant, where the mAO scores of MemTracker and MDNet drop by around 2\% when trained on imbalanced data, while the scores of GOTURN and SiamFCv2 increase by around 1\%. We assume the reason to be that the generalization ability of deep trackers is affected by several aspects of data diversity, such as the diversity in object and motion classes, scenes and challenging attributes, and the impact of object class distribution alone on the deep trackers' performance could be limited.

\noindent\textbf{Test on balanced v.s. imbalanced data.}
While we limit the maximum number of videos per class to 8 (1.9\% of the test set size) and use the class-balanced metrics, the class distribution of our test data is after all slightly imbalanced. To assess how such slight imbalance influences the evaluation stability, we randomly sample balanced test sets, where the classes are evenly distributed; and imbalanced test sets, where the class distributions are close to our test data. Then we evaluate the ranking stability on them. The results are shown in Table~\ref{tab:balanced_imbalanced_test}. We observe that, although the standard deviations of ranks assessed on the balanced test sets are relatively smaller than on the imbalanced ones, the differences are marginal (around 0.1). Besides, the average absolute differences between the method ranks on the two types of test sets are small ($1.25\sim 1.3$), indicating that the slight class-imbalance in our test set on the evaluation stability is limited under metrics mAO and mSR.

\noindent\textbf{Evaluation on seen v.s. unseen classes.} We carry out experiments to compare the performance of deep trackers on seen and unseen test data, so as to assess the generalization ability of deep trackers. Specifically, we randomly sample 4000 videos from GOT-10k for training, 240 videos of seen classes and another 240 videos of unseen classes for evaluation. Considering the randomness in sampling, we repeat the experiments three times. The results are shown in Table~\ref{tab:seen_unseen}. By comparing the tracking results on unseen test sets against seen ones, we observe an obvious performance degradation (by 0.1\%$\sim$5.9\%) for almost all trackers. Such degradation suggests the limitation of deep trackers on unfamiliar objects, and also verifies the rationality of our \textit{one-shot protocol} in promoting the development of general purposed trackers.

\noindent\textbf{Cross dataset evaluation.} We conduct experiments to compare the training performance of deep trackers on GOT-10k and commonly used ImageNet-VID datasets. Specifically, we re-train the deep trackers on GOT-10k and ImageNet-VID, and evaluate their performance on OTB2015 and GOT-10k. The results are summarized in Table~\ref{tab:cross_dataset}. On OTB2015 dataset, MemTracker and SiamFCv2 trained on GOT-10k achieve around 1\% absolute gain in AO score, compared to those trained on ImageNet-VID; while GOTURN and MDNet obtain worse results on OTB2015 when trained on GOT-10k. On the other hand, when evaluated on GOT-10k, all trackers show significant improvements (by 1.1\%$\sim$2.2\%) when trained on GOT-10k's training set, compared to ImageNet-VID. The cross-dataset evaluation results suggest the potential domain gap between OTB2015 and GOT-10k. OTB2015 contains only a few common object classes (e.g., \textit{persons, cars, faces} and \textit{bicycles}), and the scale and aspect ratio changes are small for most targets. By contrast, GOT-10k populates over 560 classes of moving objects and a variety of motion forms. Therefore, training on a lower diversity dataset (i.e., ImageNet-VID, which contains only 30 object classes) may limit the generalization ability of deep trackers on our test data.

\section{Conclusion}\label{sec:conclusion}

In this paper, we introduce GOT-10k, a large, high-diversity, and \textit{one-shot} tracking database with an unprecedentedly wide coverage of real-world moving objects. GOT-10k collects over 10,000 videos of 563 object classes and annotates 1.5 million tight bounding boxes manually. It is also the first tracking dataset that follows the \textit{one-shot protocol} to promote generalization in tracker development. We first describe the construction of GOT-10k, showing how diversity and quality are ensured in our collection and annotation stages. Then we present the principle we follow and the analytical experiments we carry out for the establishment of an efficient and relatively unbiased evaluation platform for generic purposed trackers. Finally, we train and evaluate a number of recent tracking approaches on our dataset and analyze their results. We show the major challenges of generic object tracking in real-world unconstrained scenarios and discuss the impact of training data on tracking performance. We hope GOT-10k will spawn new research in the field of generic object tracking and beyond.

\ifCLASSOPTIONcaptionsoff
  \newpage
\fi


\ifCLASSOPTIONcompsoc
  \section*{Acknowledgments}
\else
  \section*{Acknowledgment}
\fi

This work is supported in part by the National Key Research and Development Program of China (Grant No. 2016YFB1001001 and No. 2016YFB1001005), the National Natural Science Foundation of China (Grant No. 61602485 and No. 61673375), and the Projects of Chinese Academy of Science (Grant No. QYZDB-SSW-JSC006).
Lianghua~Huang and Xin~Zhao contributed equally to this work.


%

\newpage

\begin{IEEEbiography}[{\includegraphics[width=1.25in,height=1.25in,clip,keepaspectratio]{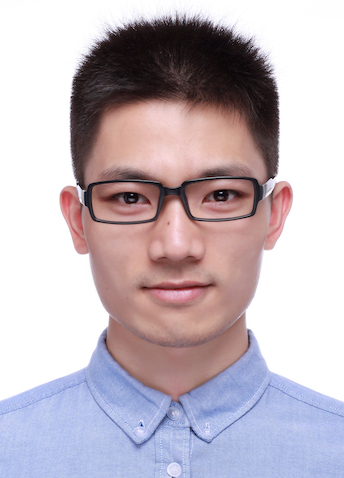}}]{Lianghua Huang} received his BSc from Dalian University of Technology (DUT) and MSc from Beijing Institute of Technology (BIT). In September 2017, he joined the Institute of Automation of the Chinese Academy of Sciences (CASIA), where he is currently studying for his doctorate. He has published 8 papers in the areas of computer vision and pattern recognition at international journals and conferences such as ICCV, AAAI, TIP, TCYB and ICME. His current research interests include pattern recognition, computer vision and machine learning.
\end{IEEEbiography}

\begin{IEEEbiography}[{\includegraphics[width=1.25in,height=1.25in,clip,keepaspectratio]{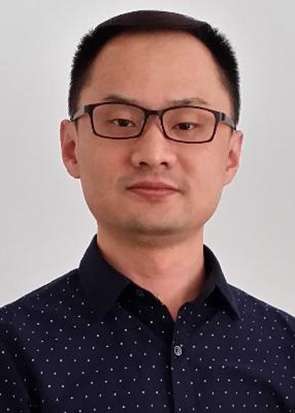}}]{Xin Zhao} received the Ph.D. degree from the University of Science and Technology of China. He is currently an Associate Professor in the Institute of Automation, Chinese Academy of Sciences (CASIA). His current research interests include pattern recognition, computer vision, and machine learning. He received the International Association of Pattern Recognition Best Student Paper Award at ACPR 2011. He received the 2nd place entry of COCO Panoptic Challenge at ECCV 2018.
\end{IEEEbiography}

\begin{IEEEbiography}[{\includegraphics[width=1.25in,height=1.25in,clip,keepaspectratio]{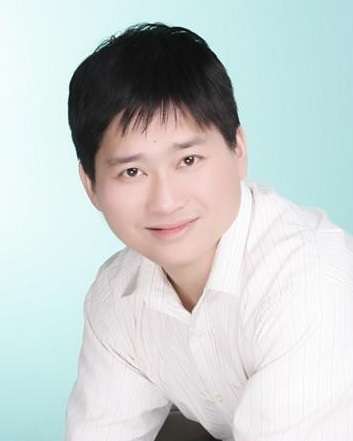}}]{Kaiqi Huang} received his BSc and MSc from Nanjing University of Science Technology, China and obtained his PhD degree from Southeast University, China. He has been a full professor in National Lab. of Pattern Recognition (NLPR), Institute of Automation, Chinese Academy of Science (CASIA) since 2005. He is IEEE Senior member and deputy general secretary of IEEE Beijing Section (2006-2008). His current researches focus on computer vision and pattern recognition including object recognition, video analysis and visual surveillance and so on. He has published over 180 papers in the important international journals and conferences such as IEEE TPAMI, T-IP, T-SMCB, TCSVT, Pattern Recognition (PR), CVIU, ICCV, ECCV, CVPR, ICIP, ICPR. He received the winner prizes of the object detection tasks in both PASCAL VOC'10 and PASCAL VOC'11, object classification task in ImageNet-ILSVRC2014, He serves as co-chairs and program committee members over forty international conferences such as ICCV, CVPR, ECCV, IEEE workshops on visual surveillance and so on. He is an executive team member of IEEE SMC Cognitive Computing Committee as well as Associate Editor of IEEE Trans on Systems, Man, and Cybernetics: Systems (TSMCs) and Pattern Recognition (PR). He has obtained some awards including CCF-IEEE Young Scientist Award (2016), the Excellent Young Scholars Award of National Science Foundation of China (NSFC) (2013), The National Science and Technology Progress Award (2011).
\end{IEEEbiography}

\end{document}